\documentclass[preprint,11pt]{elsarticle}

\usepackage{graphicx}%
\usepackage{subcaption}%
\usepackage{multirow}%
\usepackage{amsmath,amssymb,amsfonts}%
\usepackage{amsthm}%
\usepackage{mathrsfs}%
\usepackage[title]{appendix}%
\usepackage{xcolor}%
\usepackage{textcomp}%
\usepackage{manyfoot}%
\usepackage{booktabs}%
\usepackage{algorithm}%
\usepackage{algorithmicx}%
\usepackage{algpseudocode}%
\usepackage{listings}%
\usepackage{lipsum}
\usepackage{cite}
\usepackage{lmodern}
\usepackage{tabularx}
\usepackage{graphicx}
\usepackage{adjustbox}
\usepackage{rotating}
\usepackage{svg}
\usepackage{wrapfig}
\usepackage{enumitem}
\usepackage{longtable}
\usepackage{multicol}
\usepackage{lscape}
\usepackage{array}
\usepackage{soul}
\usepackage[acronym]{glossaries}

\newacronym{arima}{ARIMA}{Autoregressive Integrated Moving Average}
\newacronym{cgan}{cGAN}{Conditional Generative Adversarial Network}
\newacronym{cnn}{CNN}{Convolutional Neural Networks}
\newacronym{client}{Client}{Cross-Variable Linear Integrated Enhanced Transformer}
\newacronym{ett}{ETT}{Electricity Transformer Temperature}
\newacronym{fedformer}{FEDformer}{Frequency Enhanced Decomposed Transformer}
\newacronym{gan}{GAN}{Generative Adversarial Network}
\newacronym{mae}{MAE}{Mean Absolute Error}
\newacronym{mas}{MAS}{Multi-Agent System}
\newacronym{mse}{MSE}{Mean Square Error}
\newacronym{rnn}{RNN}{Recurrent Neural Network}
\newacronym{sota}{SOTA}{State of the Art}
\newacronym{ai}{AI}{Artificial Intelligence}
\newacronym{es}{ES}{Exponential Smoothing}
\newacronym{svm}{SVM}{Support Vector Machine}
\newacronym{snp500}{S\&P 500}{Standard and Poor's 500}
\newacronym{knn}{KNN}{K-Nearest Neighbors}
\newacronym{ann}{ANN}{Artificial Neural Network}
\newacronym{lstm}{LSTM}{Long Short-Term Memory}
\newacronym{mlp}{MLP}{Multi-Layer Perceptron}
\newacronym{uci}{UCI}{University of California, Irvine}
\newacronym{cdc}{CDC}{Centers for Disease Control and Prevention}
\newacronym{omp}{OMP}{Orthogonal Matching Pursuit}
\newacronym{patchtst}{PatchTST}{patch time series Transformer}
\newacronym{nvidia}{NVIDIA}{NVIDIA Corporation}
\newacronym{n-beats}{N-BEATS}{Neural Basis Expansion Analysis for Interpretable Time Series}
\newacronym{nasdaq}{NASDAQ}{National Association of Security Dealers Automated Quotations}
\newacronym{nyse}{NYSE}{New York Stock Exchange}
\newacronym{rtx}{RTX}{Ray Tracing Texel eXtreme}
\newacronym{gpu}{GPU}{Graphics Processing Unit}
\newacronym{gddr6}{GDDR6}{Graphics Double Data Rate 6}
\newacronym{cuda}{CUDA}{Compute Unified Device Architecture}
\newacronym{xgb}{XGB}{Extreme Gradient Boosting}
\newacronym{lgbm}{LGBM}{Light Gradient Boosting Machine}
\newacronym{}{}{}

\begin{document}
\begin{frontmatter}

\title{ForecastGAN: A Decomposition-Based Adversarial Framework for Multi-Horizon Time Series Forecasting}

\author[1]{Syeda Sitara Wishal Fatima}\corref{cor1}
\ead{fatima92@uwindsor.ca}

\author[1]{Afshin Rahimi}

\affiliation[1]{organization={Department of Mechanical, Automotive and Materials Engineering},
            addressline={401 Sunset Ave}, 
            city={Windsor},
            postcode={N9B 3P4}, 
            state={ON},
            country={Canada}}

\cortext[cor1]{Corresponding author}

\begin{abstract}
    Time series forecasting is essential across domains from finance to supply chain management. This paper introduces ForecastGAN, a novel decomposition based adversarial framework addressing limitations in existing approaches for multi-horizon predictions. Although transformer models excel in long-term forecasting, they often underperform in short-term scenarios and typically ignore categorical features. ForecastGAN operates through three integrated modules: a Decomposition Module that extracts seasonality and trend components; a Model Selection Module that identifies optimal neural network configurations based on forecasting horizon; and an Adversarial Training Module that enhances prediction robustness through Conditional Generative Adversarial Network training. Unlike conventional approaches, ForecastGAN effectively integrates both numerical and categorical features. We validate our framework on eleven benchmark multivariate time series datasets that span various forecasting horizons. The results show that ForecastGAN consistently outperforms state-of-the-art transformer models for short-term forecasting while remaining competitive for long-term horizons. This research establishes a more generalizable approach to time series forecasting that adapts to specific contexts while maintaining strong performance across diverse data characteristics without extensive hyperparameter tuning.
\end{abstract}

\begin{keyword}
     Time series forecasting \sep Generative adversarial networks \sep Time series decomposition \sep Multi-horizon prediction 
\end{keyword}

\end{frontmatter}

\section{Introduction}
\label{sec:intro}
Time series data is omnipresent in today's digital and data-abundant world. Time series forecasting serves as a critical tool in numerous applications that involve both univariate data (e.g., daily stock prices \citep{Miller2021}) and multivariate data (e.g. temperature, humidity and wind speed for weather forecasting \citep{Hewage2020}). The versatility of time series forecasting in handling these diverse data types underscores its significance in modern analytics. \par
Over the past several decades, time series forecasting has evolved significantly. Traditional statistical models such as \gls{arima} \citep{Punyapornwithaya2023} initially dominated the field. These were followed by classical machine learning techniques including Gradient Boosting \citep{Abbasimehr2023}, Random Forest \citep{randfor}, and Support Vector Machines \citep{Kurani2021}. The emergence of artificial intelligence further transformed forecasting capabilities through advanced architectures like Recurrent Neural Networks \citep{Dudukcu2023} and Convolutional Neural Networks \citep{cnn}, which capture complex non-linear patterns in time series data.
More recently, two significant developments have shaped the field: (1) the introduction of \gls{gan}s \citep{goodfellow2014generative}, which enable more robust adversarial training approaches, and (2) Transformer architectures \citep{vaswani2017attention}, which have demonstrated remarkable capabilities in sequence modeling across multiple domains. \par
Despite these advances, current models exhibit domain-specific performance characteristics that limit their generalizability. For example, transformer based models excel in long-term forecasting, but often underperform in short-term scenarios \citep{zeng2023transformers}. This performance discrepancy was highlighted in our previous comparative study \citep{WishalFatima2023}, which revealed significant variability in model performance between different datasets and forecasting horizons. These limitations indicate a need for more adaptive forecasting frameworks that can leverage the strengths of existing state-of-the-art models while maintaining flexibility across diverse forecasting contexts. Additionally, most current approaches focus exclusively on numerical features, neglecting the valuable information contained in categorical variables that are common in real-world time series data.\par
\begin{figure*}[t]
    \centering
    \includegraphics[width=\textwidth]{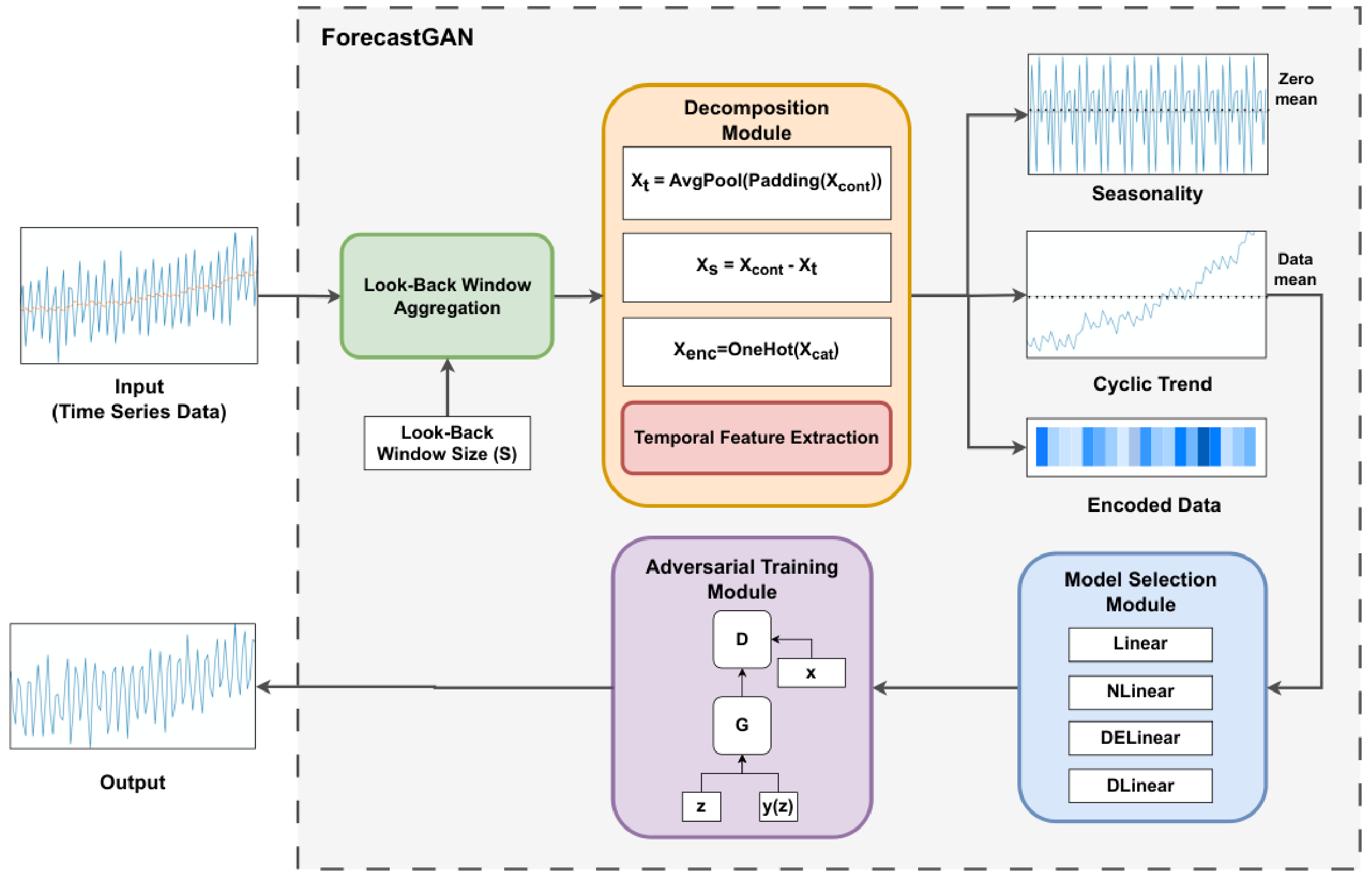}
    \caption{ForecastGAN architecture diagram (Decomposition module has the time series decomposition element, model selection module performs model selection on four of the available models, and adversarial training module is a cGAN model with a deterministically selected Generator explained in section \ref{sec:Decomp_module}, section \ref{sec:Model_module} and \ref{sec:Adversarial_module} respectively)}
    \label{fig:forecastgan}
\end{figure*}

We propose ForecastGAN, a novel modular framework that addresses these challenges through a systematic decomposition-based approach with adversarial training. Our architecture consists of three specialized, interconnected modules:
\begin{enumerate}
\item \textbf{Decomposition Module:} Extracts seasonal and trend components from numerical features while encoding categorical variables to maintain their information content.
\item \textbf{Model Selection Module:} Dynamically selects the optimal model architecture based on dataset characteristics and forecasting horizon.
\item \textbf{Adversarial Training Module:} Employs \gls{cgan} training to enhance the robustness and accuracy of predictions.
\end{enumerate}
This modular design allows each component to be optimized independently while ensuring effective communication between modules. The framework maintains abstraction between various aspects of data processing, enabling more flexible adaptation to different forecasting scenarios. The architecture is presented in Figure ~\ref{fig:forecastgan}. \par

The integration of these approaches is theoretically motivated by their complementary strengths. Time series decomposition isolates more predictable patterns (seasonality and trends), making the forecasting task more manageable. Model selection addresses the horizon-specific performance characteristics of different architectures. Finally, adversarial training transforms otherwise deterministic models into probabilistic ones, enhancing their robustness to data variability and uncertainty. From a mathematical perspective, \gls{cgan} learn the conditional probability distribution $P(X_{t+T}|X_t, ..., X_0)$ of future values given historical data. This probabilistic approach better captures the inherent uncertainty in forecasting tasks compared to deterministic point estimates, particularly when dealing with complex multivariate time series. The contributions of this paper are:
\begin{itemize}
\item A robust modular framework that delivers consistent performance across diverse forecasting horizons and datasets by separating the forecasting process into specialized functional components.
\item Empirical validation of adversarial training's effectiveness in improving predictive accuracy for otherwise deterministic forecasting models.
\item New insights into the relationship between look-back window size, and forecasting horizon, with implications for future forecasting research.
\item Comprehensive evaluation across eleven benchmark datasets demonstrating an average 37.54\% improvement over state-of-the-art transformer models for short-term forecasting while maintaining competitive performance for long-term horizons.
\end{itemize}

The remainder of this paper is organized as follows: Section \ref{sec:related_work} reviews related literature on time series forecasting methods. Section \ref{sec:background} provides theoretical background on the key concepts underlying our approach. Section \ref{sec:methodology} details the ForecastGAN architecture and its components. Section \ref{sec:experiment} describes our experimental methodology. Section \ref{sec:result} presents and discusses results. Finally, Section \ref{sec:conclusion} concludes the paper and suggests directions for future research.

\section{Related Work}
\label{sec:related_work}
The ForecastGAN architecture involves multiple concepts including time series decomposition, adversarial training, etc. We discuss the existing research to lay the foundation for model architecture. We start with discussing the model evolution for time series forecasting, followed by the applications of GANs for time series forecasting. \par

\subsection{Time Series Forecasting Models}
Traditional statistical models such as \gls{arima} and \gls{es} are widely used for industrial time-series forecasting due to their simplicity and interpretability \citep{hyndman2018}. In some cases, these models demonstrate satisfactory performance but struggle with complex datasets that exhibit nonlinear features \citep{gardner2006exponential}. Machine learning techniques such as \gls{svm}, \gls{knn}, and \gls{ann} emerge as promising alternatives, offering improved performance in capturing complex relationships and nonlinearity in time series data. However, these models often require more computational resources and can be less interpretable in comparison to traditional models \citep{mahaseth2022short}. Deep learning models currently show superior performance in various industrial forecasting tasks, including \gls{lstm} networks and \gls{cnn}s. They excel by modeling long-term dependencies and handling high-dimensional data \citep{lecun2015deep}. Nevertheless, these models can be computationally intensive and might need substantial training data for optimal performance. Hybrid models, combining different techniques, are proposed to overcome the limitations of individual models. By integrating traditional models with machine learning or deep learning techniques, hybrid models improve performance and adaptability in various industrial forecasting tasks \citep{kamarianakis2012real}. The drawback of hybrid models is they are more accurate around particular use cases and are less likely to be effective around wider conditions. While \gls{gan}s hold potential in time-series forecasting, they face challenges like training difficulty and mode collapse. Applying \gls{gan}s in time-series forecasting remains an active research area, with ongoing development of new techniques and refinements to address these challenges. A detailed discussion on time series forecasting models and their comparison is presented in one of our earlier works \citep{Fatima2024}. Furthermore, the comparison of some \gls{sota} models for long-term and short-term forecasting has been explained in detail in another paper where we have explored the strengths of some models depending upon the forecasting horizon and the chaotic element in the training data \citep{WishalFatima2023}. 

\subsection{GANs for Time Series Forecasting}
The absence of a standardized evaluation framework for \gls{gan}s initially restricted their application to fields where their outputs could be visually interpreted, such as in image generation. However, the scope of \gls{gan}s has expanded recently to include time-series data, finding applications across diverse sectors, including healthcare, finance, and the energy sector \citep{wiese2020quant}. For instance, \gls{gan}s combined with auto-regressive models have been explored for enhanced sequential data generation. Techniques such as conditioning \gls{gan}s on timestamp information have been developed to manage irregular sampling intervals. In probabilistic forecasting, conditional \gls{gan}s have been increasingly utilized. For example, Koochali et al. employed a conditional \gls{gan} integrated with \gls{lstm} units for univariate time series modeling, testing it on both synthetic and real-world datasets \citep{koochali2019probabilistic}. Another study used a Conditional \gls{gan} with \gls{lstm} and \gls{mlp} components for predicting daily stock closing prices, incorporating \gls{mse} with the generator loss to enhance performance \citep{zhang2019stock}. Zhou et al. applied \gls{lstm} and \gls{cnn} in an adversarial training framework for forecasting in the high-frequency stock market, focusing on minimizing forecast errors such as \gls{mae} or \gls{mse} alongside the \gls{gan}’s objective function \citep{zhou2018stock}.\par
Lin et al. proposed a traffic flow forecasting model sensitive to pattern variations, capable of providing accurate predictions in abnormal conditions without compromising regular performance \citep{lin2018pattern}. This model uses a \gls{cgan} with an \gls{mlp} structure and introduces two additional error terms to the standard generator loss, focusing on forecast error and reconstruction error. These advancements demonstrate the growing versatility and applicability of \gls{gan}s in time series forecasting across various sectors. Some of the popular \gls{gan}s architectures and their applications have been shown in the Appendix in Table \ref{tbl:gans_reviewed}.

\section{Background}
\label{sec:background}
This section establishes the theoretical foundations for ForecastGAN's modular architecture. We first formalize the multi-horizon time series forecasting problem, then explore the theoretical underpinnings of each core component: time series decomposition, model selection for varying horizons, and adversarial training with \gls{cgan}.

\subsection{Multi-Horizon Time Series Forecasting}
To design a multivariate forecasting model consider multivariate time-series $X = {X_0, X_1, ..., X_T}$, where each $X_t = {x_{t,1}, x_{t,2}, ..., x_{t,f}}$ represents a feature vector at time step $t$, with $f$ being the number of feature set and $x_{t,f}$ denotes the data point at time step $t$ for feature $f$. The look-back or the sliding window is the span of past time steps to make predictions. Let $S$ be the sliding window size and $T$ be the future timesteps or the forecasting horizon. Given the historical data $X = {\{X_1^t, X_2^t, ...,X_f^t\}}_{t=1}^S$ the objective for this architecture is to predict the future values $\hat{X} = {\{\hat{X}_1^t, \hat{X}_2^t, ...,\hat{X}_f^t\}}_{t=S+1}^{S+T}$ where $X_i^t$ is the value of variable $i$ at timestep $t$, $\hat{X}_i^t$ is the predicted value after $T$ timesteps. For $T =1$, the forecasting model only gives point-wise predictions rather than a future trend. For $T >1$ the forecasting model uses single-step forecasting iteratively to predict $HT$ future values where $H$ is the multiplying factor for the number of steps in predictions. This is called iterative multi-step forecasting, which is used in this paper. In iterative multi-step forecasting, the one-step prediction is made, and for the next step, this predicted value is fed back into the model. The prediction process for iterative multi-step forecast can be given by equation \ref{eq1}.
\begin{align}
\hat{X}_{t+T} &= f(X_t) \nonumber \\
\hat{X}_{t+2T} &= f([X_t, \hat{X}_{t+T}]) \nonumber \\
&\vdots \nonumber \\
\hat{X}_{t+HT} &= f([X_t, \hat{X}_{t+T}, \hat{X}_{t+2T}, \ldots, \hat{X}_{t+HT-1}]) \label{eq1}
\end{align}

The other method for predicting the next steps is direct multi-step forecasting, in which separate models are trained for each forecasting step. Each model directly predicts the value of the time series at a specific future time step. This approach can mathematically be represented as equation \ref{eq2}. 

\begin{align}
X_{t+T} &= f_1(X_t) \nonumber\\
X_{t+2T} &= f_2(X_t) \nonumber\\
&\vdots \nonumber\\
X_{t+HT} &= f_H(X_t) \label{eq2}
\end{align}

Each approach has theoretical advantages and limitations. Iterative methods can accumulate errors over multiple steps, particularly when the forecasting model has significant uncertainty. Conversely, direct methods require training multiple models, increasing computational complexity but potentially yielding higher accuracy for specific horizons. For medium to large values of $T$, direct multi-step forecasting often produces more accurate results by optimizing each model for its specific target horizon. ForecastGAN leverages this insight by employing a model selection approach that considers the specific forecasting horizon.

\subsection{Time Series Decomposition}
Harvey and Peters \citep{Harvey1990} initially presented the idea of decomposing time series data into multiple cyclic and ordered sets, proposing that the original data can be divided into trend, seasonality, and holiday components. Classical decomposition theory separates a time series into:
\begin{equation}
X_t = T_t + S_t + R_t
\end{equation}
Where $T_t$ represents trend, $S_t$ represents seasonality, and $R_t$ represents residuals or irregular components. This decomposition provides several theoretical advantages:
\begin{itemize}
\item \textbf{Complexity Reduction:} By isolating predictable patterns (trend and seasonality), the forecasting task becomes more manageable \citep{wu2021autoformer}.
\item \textbf{Component-Specific Modeling:} Different components may benefit from different modeling approaches. For instance, trend components often exhibit smoother patterns suitable for linear models, while seasonal components may require more flexible nonlinear approaches \citep{taylor2018forecasting}.
\item \textbf{Feature Enhancement:} Decomposition effectively creates new features that capture different temporal dynamics, enriching the information available to subsequent modeling stages \citep{neural_forecasting}.
\end{itemize}

Some famous examples of using decomposition as a preprocessing tool for historical data are seen in Prophet \citep{taylor2018forecasting} where the input data is divided into a trend, seasonality, and holiday components,\gls{n-beats} model \citep{nbeats} uses a similar concept in basis expansion for univariate time series point forecasting and DeepGLO \citep{NEURIPS2019_3a0844ce} uses the concept of dividing the original time series in $k$ basis time series with matrix factorization .\par

In our implementation, we employ average pooling with appropriate padding to extract trend components, following the approach in \citep{wu2021autoformer}. The trend cyclic component captures the long-term data trends, and seasonality captures the apparent effects of certain time elements on the underlying value. 
Consider the original time series as $X \in \mathbb{R}^{Nxf}$ where $N$ is the length of the series and $f$ is the number of features. The extracted trend $X_t \in \mathbb{R}^{Nxf}$ and $X_s \in \mathbb{R}^{Nxf}$ components can be given as: 
\begin{align}
    X_t &= AvgPool(Padding(X)) \label{eqn:trend}\\
    X_s &= X - X_t \label{eqn:season}
\end{align}
where the Average Pooling (\textit{AvgPool}) is used to divide the series into overlapping (or non-overlapping) regions and compute the average. This moving average operation is used to smooth the fluctuations in the data, making the series easier to predict. Padding is used to control the spatial dimensions of the series, i.e., to keep the length of the series the same as the original. 

\subsection{Theoretical Limitations of Transformers for Short-Term Forecasting}
Transformer models have demonstrated exceptional capabilities for long-term forecasting but often underperform in short-term scenarios. This limitation has a theoretical basis in the architecture's design:
\begin{itemize}
\item \textbf{Self-Attention Mechanism:} Transformers rely on self-attention mechanisms that are inherently permutation-invariant. While positional encoding attempts to preserve temporal order, some temporal information is lost, particularly for fine-grained short-term patterns \citep{zeng2023transformers, trans}.
\item \textbf{Parameter Efficiency:} Transformer models typically contain millions of parameters, which may lead to overfitting when applied to short-term forecasting with limited data points \citep{wu2021autoformer}.
\item \textbf{Context Window Utilization:} For short-term forecasting, local patterns within a small temporal neighborhood often contain most of the relevant information. Transformers' global attention mechanisms may unnecessarily distribute focus across the entire sequence \citep{nguyen2022improving, NEURIPS2022_266983d0}.
\end{itemize}
These theoretical considerations suggest that simpler models, such as linear networks with appropriate embeddings, might outperform transformers for short-term forecasting tasks \citep{zeng2023transformers}. This insight motivates our Model Selection Module, which can adaptively choose between different model architectures based on the forecasting horizon.

\subsection{Adversarial Training for Robust Forecasting}
Adversarial training offers a theoretical framework for enhancing model robustness by exposing the model to challenging examples during training \citep{Zhao2022}. In the context of time series forecasting, this approach addresses several fundamental challenges. Time series data often exhibits distribution shifts between training and testing periods. Adversarial training helps models become more robust to such shifts. Deterministic forecasting models provide point estimates without capturing prediction uncertainty. Adversarial frameworks, particularly GANs, learn the conditional distribution of future values, inherently capturing uncertainty. In time series with multiple possible futures, standard forecasting models might average across possibilities, producing unrealistic predictions. GANs can potentially capture multimodal future distributions.

\begin{figure*}[t]
    \centering
    \includegraphics[width=\textwidth]{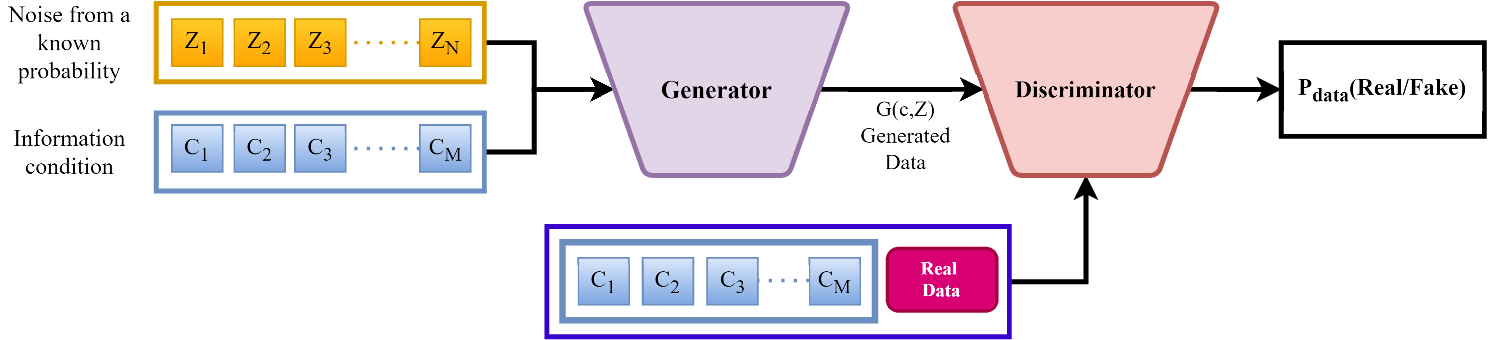}
    \caption{Structure of cGAN}
    \label{fig:cGAN}
\end{figure*}

Adversarial training is a technique employed to improve the generalization and robustness of models against adversarial attacks \citep{Zhao2022}. It involves training a model with clean and adversarially perturbed examples to make it more robust to small but intentionally worst-case perturbations. Consider a predicting model $f_\theta$ with parameters $\theta$ and input $X$ and $Y$ as the ground truth. The adversarial example $X'$ is generated by adding a small perturbation $\lambda$ to $X$ such that $X'=X+\lambda$ and $\lambda$ is designed to maximize the loss $\mathcal{L}(f_\theta (X'),Y)$. Thus, the objective in adversarial training for a data distribution $\mathcal{D}$ reduces to a min-max optimization task:
\begin{align}
\text{min}_\theta \mathbb{E}_{(X,Y)\sim\mathcal{D}}[ \text{max}_\lambda \mathcal{L}(f_\theta(X+\lambda),Y)]
\end{align}

Transitioning from deterministic to probabilistic models can further enhance the robustness of the predictive models \citep{Koochali2021}. In a deterministic model, the output $f_\theta(x)$ is a single point estimate, but a probabilistic model predicts a distribution over possible outcomes. This shift can be achieved by modeling the output as a random variable and using Bayesian methods or variational inference\citep{terzi2021adversarial}. Mathematically, for the predicted output $\hat{Y}$ instead of predicting $\hat{Y}=f_\theta(X)$, a probabilistic model predicts $\hat{Y} \sim P(Y|X, \theta)$ which is a probability distribution parameterized by $\theta$. The architecture for \gls{cgan} is shown in Fig. \ref{fig:cGAN}. 

The generator functions as the probabilistic model, while the discriminator provides essential gradients for optimizing the generator during its training phase. To learn $P(X_{t+T}|X_t, ..., X_0)$, we utilize historical data ${X_t, ..., X_0}$ as the condition in the \gls{cgan}. The generator is tasked with producing $X_{t+T}$, thereby learning the probability distribution equivalent to $P(X_{t+T}|X_t, ..., X_0)$, which is the desired target distribution. The value function employed in the training of \gls{gan} (the probabilistic forecast model) is formulated as follows:
\begin{align}
\min_{G} \max_{D} V(D, G) = \mathbb{E}_{x\sim P_{\text{data}}(x)}[\log(D(x))] + \mathbb{E}_{z\sim P_{\text{noise}}(z)}[\log(1 - D(G(z)))]
\end{align}
where $min_{G}$ $max_{D}$ represents the min-max game between the generator $G$ and discriminator $D$. $V(D, G)$ is the value function for the \gls{gan}. $log(D(x))$ is the logarithm of the probability that $D$ assigns to real data, where $x$ is the real data. $log(1 - D(G(z)))$ is the logarithm of the probability that $D$ assigns to fake data where $G(z)$ is the real data.

This probabilistic framework enables the model to quantify uncertainty in its predictions, which can be particularly valuable in adversarial settings. Training the model to predict distributions rather than point estimates makes it more adept at handling the variability and uncertainty introduced by adversarial perturbations, ultimately leading to more robust and reliable forecasting systems.

\section{Methodology}
\label{sec:methodology}
The ForecastGAN architecture has been presented in Figure \ref{fig:forecastgan}. This section presents the ForecastGAN architecture in detail. We begin with an overview of the framework, followed by in-depth explanations of each module, their interactions, and the overall workflow. The framework consists of three specialized, interconnected modules designed to address specific aspects of the forecasting challenge: 
\begin{enumerate}
\item \textbf{Decomposition Module:} Processes raw time series data by decomposing numerical features into seasonal and trend components, encoding categorical features, and extracting temporal features from date-time columns.
\item \textbf{Model Selection Module:} Evaluates multiple model architectures on the processed data to identify the optimal configuration for the specific dataset and forecasting horizon.
\item \textbf{Adversarial Training Module:} Employs conditional GAN training to enhance the robustness and accuracy of the selected model.
\end{enumerate}

This modular design enables each component to be optimized independently while maintaining effective information flow between stages. The framework supports both short-term and long-term forecasting by adaptively selecting appropriate model configurations based on the specific forecasting task.

\subsection{Look-Back Window Aggregator}
The aggregator block outputs the data according to the set look-back window size. This represents the extent of consolidating the past data and is indicative of how much micro-level information is needed for the said prediction step. For example, for single-step prediction i.e., $T=1$ a value of $S=96$ can lose information necessary for good predictions. The impact of the sliding window size is discussed in detail in Section \ref{sec:result}. For the aggregator, mean is used for the continuous variables and mode is used for the categorical variable values.  

\subsection{Decomposition Module}
\label{sec:Decomp_module}

\begin{figure}[t]
    \centering
    \includegraphics[width=\textwidth]{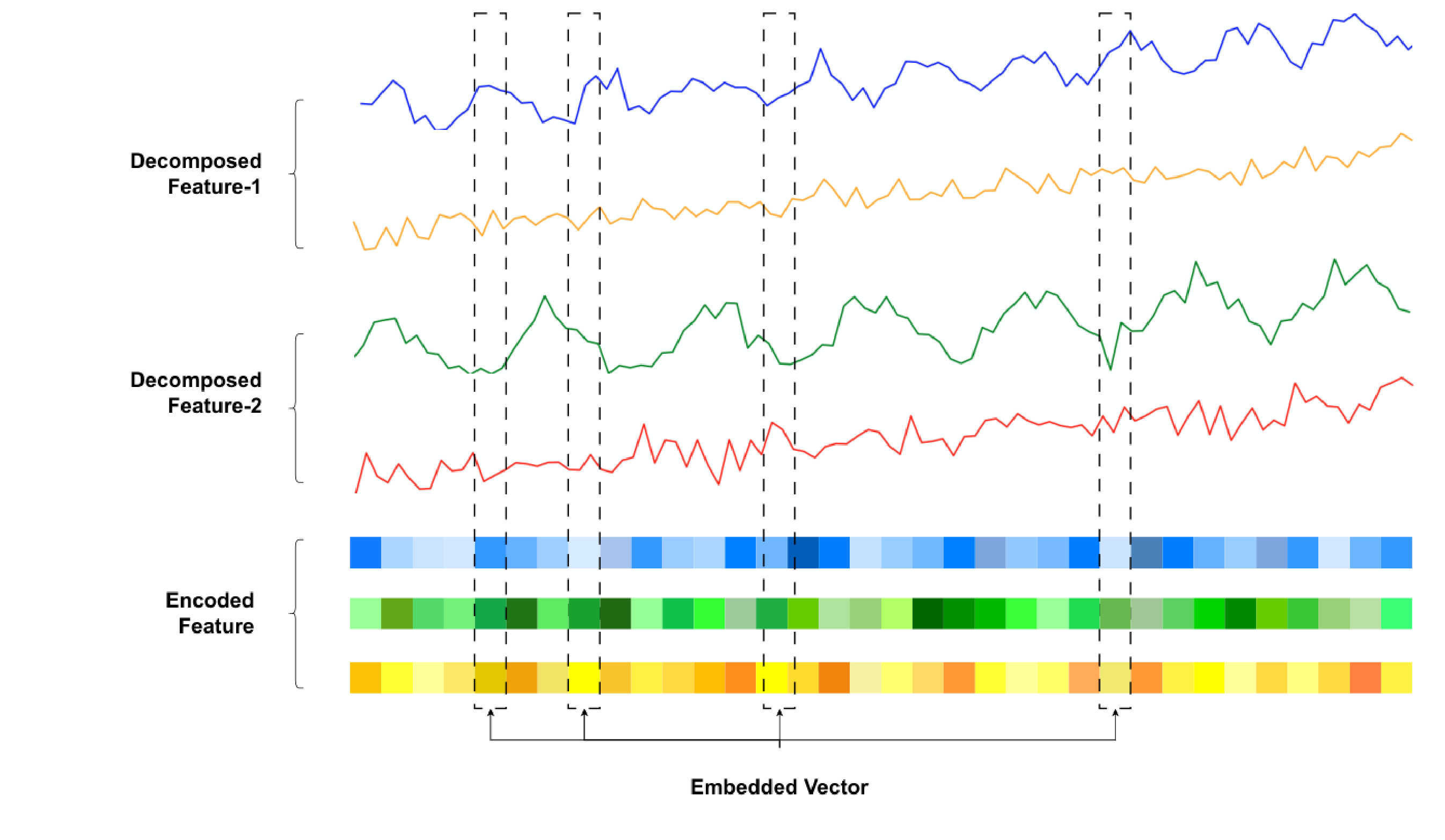}
    \caption{Embedding method for Decomp-Agent: Each of the continuous features is decomposed in trend and seasonality components and the categorical features are encoded (one-hot) whereas the dotted block represents the values of these features at the same time step embedded into a vector}
    \label{fig:embedding}
\end{figure}

The Decomposition Module serves as the preprocessing foundation for the forecasting architecture. It transforms raw multivariate time series data into a format that highlights relevant patterns and preserves the information content of different feature types. For each numerical feature, the module performs time series decomposition using the following approach:
\begin{itemize}
\item \textbf{Trend Extraction:} Apply average pooling with appropriate padding to extract the cyclic-trend component:
\begin{equation}
X_t = \text{AvgPool}(\text{Padding}(X))
\end{equation}
\item \textbf{Seasonality Extraction:} Subtract the trend component from the original series to obtain the seasonal component:
\begin{equation}
    X_s = X - X_t
\end{equation}
\end{itemize}
This decomposition isolates predictable patterns (trend and seasonality), making the forecasting task more manageable for subsequent modules.

\subsubsection{Categorical Feature Processing}
For categorical features, the module applies one-hot encoding to transform them into a numerical representation while preserving their information content. This encoding creates a binary vector for each categorical value, allowing the model to leverage categorical information without imposing arbitrary ordinal relationships.

\subsubsection{Temporal Feature Extraction}
For datetime columns, the module extracts temporal proxy features that capture cyclical patterns at different timescales:
\begin{itemize}
\item Day of week (captures weekly patterns)
\item Day of month (captures monthly patterns)
\item Month of year (captures yearly patterns)
\item Hour of day (captures daily patterns)
\item Minute of hour (captures hourly patterns)
\item Quarter (captures quarterly patterns)
\end{itemize}
These derived features provide explicit temporal context that helps models identify recurring patterns at different timescales.

\subsubsection{Feature Embedding}
The processed individual features are combined into a unified dataset and embedding is applied to preserve temporal relationships. As illustrated in Figure \ref{fig:embedding}, this embedding process creates fixed-length vectors that incorporate information from all feature types at each time step. The complete algorithm for the Decomposition Module is presented in Algorithm \ref{alg:decomp_module}. 

\begin{algorithm}[htbp!]
\caption{Decomposition Module Algorithm}
\label{alg:decomp_module}
\begin{algorithmic}[1]
\Require Multivariate time series data $X$ with $t$ time steps and $f$ features
\Ensure Decomposed, processed and embedded time series data
\State Initialize empty lists for seasonal components, trend components, and encoded categorical features
\If {data contains categorical features}
\For{each feature $f$ in $X$}
\If{feature $f$ is numerical}
\State Apply average pooling with padding on $X(f)$ to obtain the cyclic-trend component $X_t(f)$
\State Calculate the seasonality component $X_s(f)$ by subtracting $X_t(f)$ from $X(f)$
\State Append $X_t(f)$ and $X_s(f)$ to their respective lists
\ElsIf{feature $f$ is categorical}
\State Apply one-hot encoding to $X(f)$ to obtain encoded features
\State Append encoded features to categorical features list
\EndIf
\State Extract column with type datetime and generate time features
\State Combine all processed features into a single dataset
\State Embed the combined data to preserve temporal information
\State Forward the embedded data
\EndFor
\Else
\For{each feature $f$ in $X$}
\State Apply average pooling with padding on $X(f)$ to obtain the cyclic-trend component $X_t(f)$
\State Calculate the seasonality component $X_s(f)$ by subtracting $X_t(f)$ from $X(f)$
\State Append $X_t(f)$ and $X_s(f)$ to their respective lists
\State Forward the trend and seasonality list
\EndFor
\EndIf
\end{algorithmic}
\end{algorithm}

\subsection{Model Selection Module}
\label{sec:Model_module}
The Model Selection Module identifies the optimal model architecture for a given dataset and forecasting horizon. This module addresses the observation that different model architectures exhibit varying performance characteristics depending on the specific forecasting task.
\subsubsection{Model Variants}
Inspired by \citep{zeng2023transformers}, the module evaluates four variations of linear networks:
\begin{enumerate}
\item \textbf{Linear:} Simple one-layer linear model serving as a baseline. It applies a linear transformation to the original multivariate time series data:
\begin{equation}
\hat{X}_i = \mathbb{W} X_i
\end{equation}
where $\mathbb{W} \in \mathbb{R}^{T \times L}$ is the weight matrix, $X_i$ is the input for the $i$-th variable, and $\hat{X}_i$ is the corresponding prediction.
\item \textbf{NLinear:} Extends the linear model with input sequence normalization. It normalizes by subtracting the last value of the sequence from the input, applies the linear transformation, and then adds back the subtracted value:
\begin{equation}
    \hat{X}_i = \mathbb{W}(X_i - X_t) + X_t
\end{equation}
where $X_t$ is the last value in the input sequence.
\item \textbf{DELinear:} Applies a linear layer to the decomposed and embedded data for the input data containing both categorical and continuous features:
\begin{equation}
    \hat{X}_i = \mathbb{W}D_i
\end{equation}
where $D_i$ represents the decomposed and embedded data.
\item \textbf{DLinear:} For datasets without categorical features, this model applies separate linear layers to the seasonal and trend components:
\begin{align}
    \hat{X}_{s,i} &= \mathbb{W}_s X_{s,i} \\
    \hat{X}_{tr,i} &= \mathbb{W}_{tr} X_{tr,i} \\
    \hat{X}_i &= \hat{X}_{s,i} + \hat{X}_{tr,i}
\end{align}
where $X_{s,i}$ and $X_{tr,i}$ are the seasonal and trend components, respectively, and $\mathbb{W}_s$ and $\mathbb{W}_{tr}$ are their corresponding weight matrices.
\end{enumerate}
The choice of linear models is motivated by their simplicity, stability, and computational efficiency, which make them particularly well-suited for adversarial training. Additionally, recent research has shown that these models can outperform complex transformer architectures for certain forecasting tasks \citep{zeng2023transformers}.
\subsubsection{Selection Process}
The module evaluates each model on the validation set and selects the one with the lowest validation loss. This selection process can be formalized as:
\begin{equation}
M^* = \arg\min_{M \in \mathcal{M}} \mathcal{L}(M, X_{val}, Y_{val})
\end{equation}
where $M^*$ is the selected model, $\mathcal{M}$ is the set of candidate models, $\mathcal{L}$ is the loss function (e.g., MSE), and $X_{val}$ and $Y_{val}$ are the validation inputs and targets, respectively.
The complete algorithm for the Model Selection Module is presented in Algorithm \ref{alg:model_module}.

\begin{algorithm}[htbp!]
\caption{Model Selection Module Algorithm}
\label{alg:model_module}
\begin{algorithmic}[1]
\Require Input time series data $X$, decomposed datasets $X_s$ and $X_t$ from Decomposition Module, validation data
\Ensure Best linear model and corresponding predictions $\hat{X}$
\State Initialize best model as None and best loss as $\infty$
\For{each model in models}
    \If{model is Linear}
        \State Apply linear regression on $X$ using the weight vector
        \State $\hat{X}_i = \mathbb{W} X_i$
    \ElsIf{model is NLinear}
        \State Normalize $X$ by subtracting the last value of $X_t$ from each $X_i$
        \State Apply linear regression on normalized $X$
        \State $\hat{X}_i = \mathbb{W} (X_i - X_t) + X_t$
    \ElsIf{model is DELinear}
        \State Linear regression on decomposed data $D$ using the weight vector
        \State $\hat{X}_i = \mathbb{W} D_i$
    \ElsIf{model is DLinear}
        \State Apply linear regression on $X_s$ and $X_{tr}$ components separately
        \State $\hat{X}_{s,i} = \mathbb{W}_s X_{s,i}$
        \State $\hat{X}_{tr,i} = \mathbb{W}_{tr} X_{tr,i}$
        \State $\hat{X}_i = \hat{X}_{s,i} + \hat{X}_{tr,i}$
    \EndIf
    \State Calculate validation loss for the model
    \If{current loss $<$ best loss}
        \State Update best model and best loss
    \EndIf
\EndFor
\State Return best model and configuration
\end{algorithmic}
\end{algorithm}

\subsection{Adversarial Training Module}
\label{sec:Adversarial_module}
The Adversarial Training Module enhances the selected model through conditional GAN training. This approach transforms the deterministic forecasting model into a probabilistic one, improving its robustness and generalization capabilities. The module consists of two primary components: The Generator is the best model selected by the Model Selection Module serves as the generator. It takes historical time series data as input and generates future predictions. While the Discriminator is a neural network that distinguishes between real and generated time series data. The discriminator architecture includes the input layer accepting the concatenated time series data and conditional information. Hidden layers have Dense layers with LeakyReLU activation, batch normalization, and dropout for regularization. The output layer is a single unit with sigmoid activation that outputs the probability of the input being real. The adversarial training process involves two alternating steps:
\begin{itemize}
\item \textbf{Discriminator Training:}
\begin{itemize}
\item Sample real data from the training set
\item Generate fake data using the generator
\item Compute discriminator loss for real data:
\begin{equation}
\mathcal{L}{\text{real}} = \text{BCE}(D(X{\text{real}}|c), 1)
\end{equation}
\item Compute discriminator loss for fake data:
\begin{equation}
\mathcal{L}{\text{fake}} = \text{BCE}(D(G(z|c)), 0)
\end{equation}
\item Update discriminator parameters to minimize the combined loss:
\begin{equation}
\mathcal{L}D = \mathcal{L}{\text{real}} + \mathcal{L}{\text{fake}}
\end{equation}
\end{itemize}
\item \textbf{Generator Training:}
\begin{itemize}
    \item Generate fake data using the generator
    \item Compute adversarial loss to fool the discriminator:
    \begin{equation}
        \mathcal{L}_G = \text{BCE}(D(G(z|c)), 1)
    \end{equation}
    \item Update generator parameters to minimize the adversarial loss
\end{itemize}
\end{itemize}
where BCE represents binary cross-entropy loss, $D$ is the discriminator, $G$ is the generator, $z$ is random noise, and $c$ is the conditional information (historical time series data).

\subsubsection{GAN Stability Measures}
GAN training is notoriously unstable, especially for time series data. We implement several measures to enhance stability:
\begin{itemize}
\item \textbf{Gradient Penalty:} We apply a gradient penalty to the discriminator's loss to enforce Lipschitz continuity, which helps prevent mode collapse and gradient explosion:
\begin{equation}
\mathcal{L}{GP} = \lambda{GP} \mathbb{E}{\hat{x} \sim \mathbb{P}{\hat{x}}}[(|\nabla_{\hat{x}}D(\hat{x})|2 - 1)^2]
\end{equation}
where $\hat{x}$ is a sample from a distribution $\mathbb{P}{\hat{x}}$ that interpolates between real and generated samples.
\item \textbf{Spectral Normalization:} Applied to the discriminator's weights to constrain its Lipschitz constant, further stabilizing training.
\item \textbf{Two-Timescale Update Rule (TTUR):} Different learning rates for the generator and discriminator, which has been shown to improve convergence.
\end{itemize}

\subsubsection{Inference Process}
For inference (making predictions on new data), we use only the generator component of the trained GAN. The generator takes historical time series data as input and produces forecasts for the specified horizon.
The complete algorithm for the Adversarial Training Module is presented in Algorithm \ref{alg:adversarial_module}.

\begin{algorithm}[htbp!]
\caption{Adversarial Training Module Algorithm}
\label{alg:adversarial_module}
\begin{algorithmic}[1]
\Require Best model parameters from Model Selection Module, training data $X_{\text{train}}$, test data $X_{\text{test}}$, number of epochs
\Ensure Adversarially trained generator model
\State Initialize generator using the best model architecture and weights from Model Selection Module
\State Initialize discriminator with neural network architecture including batch normalization and dropout
\For{specified number of epochs}
    \State \textbf{Step 1: Train Discriminator}
    \begin{itemize}
        \item Sample batch of real data $X_{\text{real}}$ from $X_{\text{train}}$
        \item Generate batch of fake data $X_{\text{fake}} = \text{Generator}(X_{\text{input}})$
        \item Compute discriminator loss for real data:
        \item $\mathcal{L}_{\text{real}} = \text{BCE}(\text{Discriminator}(X_{\text{real}}), 1)$
        \item Compute discriminator loss for fake data:
        \item $\mathcal{L}_{\text{fake}} = \text{BCE}(\text{Discriminator}(X_{\text{fake}}), 0)$
        \item Apply gradient penalty
        \item Combine losses and update discriminator parameters:
        \item $\mathcal{L}_{\text{D}} = \mathcal{L}_{\text{real}} + \mathcal{L}_{\text{fake}} + \lambda_{\text{GP}}\mathcal{L}_{\text{GP}}$
        \item Update discriminator parameters to minimize $\mathcal{L}_{\text{D}}$
    \end{itemize}
    
    \State \textbf{Step 2: Train Generator Adversarially}
    \begin{itemize}
        \item Sample batch of input data $X_{\text{input}}$ from $X_{\text{train}}$
        \item Generate batch of fake data $X_{\text{fake}} = \text{Generator}(X_{\text{input}})$
        \item Compute generator loss to fool the discriminator:
        \item $\mathcal{L}_{\text{G}} = \text{BCE}(\text{Discriminator}(X_{\text{fake}}), 1)$
        \item Update generator parameters to minimize the adversarial loss $\mathcal{L}_{\text{G}}$
    \end{itemize}
\EndFor
\State Return the adversarially trained generator model
\end{algorithmic}
\end{algorithm}

\subsection{Complexity Analysis}
The computational complexity of ForecastGAN can be analyzed for each module:
For the Decomposition Module, the complexity is dominated by the average pooling operation, which has a complexity of $O(nf)$, where $n$ is the number of time steps and $f$ is the number of features. In Model Selection Module, the linear models have training complexity of $O(nfd)$, where $d$ is the dimensionality of the feature space after decomposition and embedding. Evaluating all four model variants has a complexity of $O(4nfd)$. Lastly, for the Adversarial Training Module, the complexity depends on the selected model architecture and the number of training epochs. For a linear generator, the complexity is approximately $O(enfd)$, where $e$ is the number of epochs.\par
The overall computational complexity of ForecastGAN is therefore $O(nf + 4nfd + enfd) = O(nfd(4 + e))$, which is significantly lower than transformer-based approaches with complexity on the order of $O(n^2d)$ due to the self-attention mechanisms \citep{zeng2023transformers}. In practice, this translates to faster training times. For example, on the ETTh1 dataset with $T=96$, ForecastGAN trains in approximately 15 minutes on a single NVIDIA RTX GPU, compared to over an hour for transformer-based models like Informer on the same hardware \citep{haoyietal-informer-2021}.

\section{Experiments}
\label{sec:experiment}
To evaluate ForecastGAN comprehensively, we conducted extensive experiments across multiple datasets with varying forecasting horizons. This section details our experimental methodology, including datasets, baseline models, evaluation metrics, and implementation details.

\subsection{Datasets} 
\label{sec:dataset}
Extensive experiments are conducted for eleven standard real-world multivariate time series datasets for long-term forecasting. The complete details of these datasets are given in \ref{tbl:datasets}. 
\begin{table}[htbp!]
\centering
\caption{Details of eleven popular multivariate time series datasets used for ForecastGAN evaluation}
\begin{tabular}{cccc}
\hline
\textbf{Dataset} & \textbf{Features} & \textbf{Timesteps} & \textbf{Sample Rate} \\
\hline
ETTh1 & 7 & 17,420 & 1 hour \\
ETTh2 & 7 & 17,420 & 1 hour \\
ETTm1 & 7 & 69,680 & 5 minutes \\
ETTm2 & 7 & 69,680 & 5 minutes\\
Productivity & 15 & 1,197 & 1 hour \\
Electricity & 321 & 26,304 & 1 hour \\
Illness & 7 & 966 & 1 week \\
Traffic & 862 & 17,544 & 1 hour \\
Weather & 21 & 52,696 & 10 minutes \\
Exchange Rate & 8 & 7,588 & 1 day \\
Stock Price & 84 & 7,936 & 1 day \\
\hline
\end{tabular}
\label{tbl:datasets}
\end{table}

The datasets represent a wide range of forecasting challenges:
\begin{itemize}
\item \textbf{ETT (Electricity Transformer Temperature):} Four datasets (ETTh1, ETTh2, ETTm1, ETTm2) containing power load and oil temperature readings at different temporal resolutions. These datasets are widely used benchmarks for long-term forecasting \citep{haoyietal-informer-2021}.
\item \textbf{Productivity:} Records garment employee productivity measured hourly during 9-hour daily shifts. The target metric is the normalized productivity value between 0 and 1. This dataset contains both numerical and categorical features, making it particularly suitable for evaluating our framework's ability to handle mixed feature types \citep{data_prod}.
\item \textbf{Electricity:} Contains hourly electricity consumption measurements for 321 customers. This high-dimensional dataset tests the framework's scalability to large feature spaces \citep{misc_electricityloaddiagrams20112014_321}.
\item \textbf{Illness:} Weekly records of patients with flu-like illnesses from the CDC, featuring strong seasonal patterns and challenging long-term dependencies \citep{illness_data}.
\item \textbf{Traffic:} Hourly road occupancy rates measured by sensors on San Francisco Bay Area freeways. With 862 features, this is the highest-dimensional dataset in our evaluation \citep{exchange_rate_data}.
\item \textbf{Weather:} Weather condition measurements in Germany for 2020, featuring diverse meteorological variables with complex interdependencies \citep{weather_data}.
\item \textbf{Exchange Rate:} Daily exchange rates for 8 countries, characterized by high volatility and non-stationarity \citep{exchange_rate_data}.
\item \textbf{Stock Price:} Daily closing prices of major stock indices including S\&P 500, NASDAQ, Dow Jones, Russell 2000, and NYSE Composite from 2010 to 2017 \citep{stock_data}.
\end{itemize}
These datasets were selected to represent a diverse range of forecasting challenges, including different temporal resolutions (from 5 minutes to 1 week), dimensionality (from 7 to 862 features), domains (energy, transportation, health, finance, etc.), and temporal characteristics (seasonal patterns, trends, volatility, etc.).

\subsection{Data Preprocessing and Splitting}
For each dataset, we applied the following preprocessing steps:
\begin{enumerate}
\item \textbf{Missing Value Handling:} Missing values were imputed using forward fill followed by backward fill to ensure completeness.
\item \textbf{Normalization:} Numerical features were normalized using min-max scaling to the range [0,1] to ensure consistent scale across features.

\item \textbf{Train-Validation-Test Split:} Each dataset was divided into training (70\%), validation (10\%), and testing (20\%) sets using temporal splits rather than random sampling to preserve the chronological order of observations. This approach ensures that future data is not used to predict past events, maintaining the integrity of the forecasting task.
\end{enumerate}
Figure \ref{fig:data_distribution} in the Appendix illustrates the data distributions across train and test sets for all datasets, highlighting the differences in distribution that make certain datasets particularly challenging.
\subsection{Forecasting Horizons}
To evaluate performance across different forecasting scenarios, we conducted experiments with multiple prediction horizons:
\begin{itemize}
\item \textbf{Long-term Forecasting:} Horizons of $T \in \{96, 192, 336, 720\}$ time steps for most datasets, with $T \in \{24, 36, 48, 60\}$ for the Illness dataset due to its weekly sampling rate.
\item \textbf{Short-term Forecasting:} Horizons of $T \in \{12, 24, 32, 48\}$ time steps for most datasets, with $T \in \{2, 6, 8, 10\}$ for the Illness dataset.
\item \textbf{Single-step Forecasting:} Horizon of $T = 1$ to evaluate immediate next-step prediction performance.
\end{itemize}
For each forecasting horizon, we experimented with different look-back window sizes to identify optimal configurations. The primary look-back window sizes used were $S = 96$ for long-term forecasting, $S = 12$ for short-term forecasting, and $S = 1$ for single-step forecasting, with adjustments for the Illness dataset ($S = 24$, $S = 2$, and $S = 1$ respectively).

\subsection{Baseline Models}
We compared ForecastGAN against two groups of baseline models:
\subsubsection{Transformer-based Models}
For multi-step forecasting, we compared against state-of-the-art transformer models:
\begin{itemize}
\item \textbf{Informer} \citep{haoyietal-informer-2021}: A transformer model with ProbSparse self-attention that reduces complexity from $O(L^2)$ to $O(L \log L)$.
\item \textbf{Robformer} \citep{robformer}: A robust transformer architecture that integrates adaptive normalization techniques and specialized attention mechanisms designed to handle noise and outliers in time series data, resulting in improved stability for financial and volatile datasets.
\item \textbf{TimeXer} \citep{timexer}: Employs a time-frequency dual-domain modeling approach that leverages wavelet transforms to capture multi-scale temporal dynamics, particularly effective for time series with complex non-stationary behaviors.
\item \textbf{Crossformer} \citep{zhang2022crossformer}: Employs a two-stage attention mechanism to capture both temporal and feature dependencies.
\item \textbf{Pathformer} \citep{pathformer}: Introduces a path-dependent attention mechanism that models sequential dependencies through learnable routing paths, allowing the model to focus on the most relevant historical patterns for different forecasting contexts.
\item \textbf{Client} \citep{client}: Incorporates latent interval transformations to capture time series dynamics more effectively.
\end{itemize}
A comparison with some other popular transformer-based architectures including Autoformer, FEDformer and PatchTST has been provided in appendix. 

\subsubsection{Machine Learning Models for Short-term Forecasting Baseline}
For single-step forecasting, we additionally compared against traditional machine learning models, including linear approaches (Linear Regression, Bayesian Ridge Regression, Orthogonal Matching Pursuit, Huber Regressor) and tree-based ensemble methods (XGBoost, LightGBM, CatBoost, and Random Forest). These models were selected for their established performance in time series forecasting and to provide a diverse baseline spanning different algorithmic families.

\subsection{System Information}
\label{sec:append_sys_info}
For comparative evaluation purposes, baseline results for Informer, Robformer, TimeXer and Pathformer were sourced from the comprehensive benchmarking study by \citep{zeng2023transformers} and their original papers \citep{robformer, timexer, pathformer}. Results for Crossformer were partially obtained from the original publication, while additional evaluations—specifically for ETTm2, ETTh2, and alternative forecasting windows across other datasets—were independently reproduced using the official implementation available in the authors' repository\footnote{https://github.com/Thinklab-SJTU/Crossformer}. All ForecastGAN experiments and additional baseline evaluations were conducted on a high-performance computing environment equipped with dual NVIDIA Titan RTX GPUs (24GB GDDR6 memory each), utilizing CUDA 12.2 to optimize GPU acceleration and parallel processing capabilities.

\subsection{Evaluation Metrics}
We evaluated model performance using two standard metrics for regression tasks:
\begin{enumerate}
\item \textbf{Mean Absolute Error (MAE):} Measures the average absolute difference between predictions and ground truth:
\begin{equation}
\text{MAE} = \frac{1}{H} \sum_{i=1}^{H} |y_{T+i} - \hat{y}_{T+i}|
\end{equation}
\item \textbf{Mean Squared Error (MSE):} Measures the average squared difference between predictions and ground truth:
\begin{equation}
    \text{MSE} = \frac{1}{H} \sum_{i=1}^{H} (y_{T+i} - \hat{y}_{T+i})^2
\end{equation}
\end{enumerate}
where $H$ is the forecast horizon, $T$ is the length of the look-back window, $y$ is the ground truth, and $\hat{y}$ is the predicted value.
These metrics were chosen for their interpretability and compatibility with previous forecasting literature, enabling direct comparisons with state-of-the-art approaches.

\subsubsection{ForecastGAN Configuration}
The implementation details for ForecastGAN are presented in Table \ref{tbl:forecastgan_config}, which outlines the key parameters for each module.

\begin{table}[ht]
\centering
\small  
\caption{ForecastGAN implementation configuration by module}
\label{tbl:forecastgan_config}
\setlength{\tabcolsep}{4pt}  
\begin{tabular}{|p{0.2\textwidth}|p{0.25\textwidth}|p{0.45\textwidth}|}
\hline
\textbf{Module} & \textbf{Parameter} & \textbf{Configuration} \\
\hline
\multirow{3}{=}{Decomposition Module} 
& Pooling & Average pooling, kernel size 25 \\
\cline{2-3}
& Padding & 'same' (maintains temporal dimensions) \\
\cline{2-3}
& Categorical encoding & One-hot (if $\leq$10 unique values), otherwise ordinal \\
\hline
\multirow{4}{=}{Model Selection Module} 
& Training & 100 epochs with early stopping (patience=10) \\
\cline{2-3}
& Optimizer & Adam (learning rate=0.001) \\
\cline{2-3}
& Loss function & Mean Squared Error (MSE) \\
\cline{2-3}
& Batch size & 32 \\
\hline
\multirow{8}{=}{Adversarial Training Module} 
& Discriminator architecture & 3-layer MLP (128, 64 units) with LeakyReLU(0.2) \\
\cline{2-3}
& Batch normalization & Applied after each layer (momentum=0.8) \\
\cline{2-3}
& Dropout & Rate of 0.3 for regularization \\
\cline{2-3}
& Generator optimizer & Adam (lr=0.0002, $\beta_1$=0.5, $\beta_2$=0.999) \\
\cline{2-3}
& Discriminator optimizer & Adam (lr=0.0001, $\beta_1$=0.5, $\beta_2$=0.999) \\
\cline{2-3}
& Gradient penalty & $\lambda_{GP}=10$ \\
\cline{2-3}
& Training & 200 epochs with validation-based early stopping \\
\cline{2-3}
& Batch size & 64 \\
\hline
\end{tabular}
\end{table}

This modular configuration enabled efficient training while maintaining robust performance across diverse forecasting scenarios. The differential learning rates and optimization parameters between the generator and discriminator were specifically tuned to enhance GAN training stability.

\section{Results and Discussions}
\label{sec:result}
This section presents and analyzes the experimental results, comparing ForecastGAN against baseline models across different forecasting horizons and datasets. We also include results from ablation studies and sensitivity analyses to provide deeper insights into the framework's behavior.

\subsection{Long-term Forecasting Performance}
For long-term forecasting, ForecastGAN is compared against six state-of-the-art models including both transformer-based architectures (Informer, Crossformer, TimeXer) and specialized time series models (Robformer, Pathformer, Client). This comprehensive comparison is justified as these models employ direct multi-step forecasting rather than iterative approaches, which are known to suffer from error accumulation over longer horizons. Table \ref{tbl:long_results} presents the comparative results across nine benchmark datasets with varying forecasting horizons. Results marked with $*$ indicate values that were evaluated using publicly available repositories, while other baseline results were obtained from previously published benchmarks \citep{zeng2023transformers, client}. ForecastGAN demonstrates strong performance across multiple datasets, with the most substantial improvements observed on the Exchange Rate dataset (average improvement of 26.73\% across all horizons) and ETTm1 dataset (average improvement of 10.77\%). The results show particular strength in capturing complex patterns for datasets with pronounced seasonality and trend components. Conversely, more modest performance is observed for the Traffic dataset, where ForecastGAN shows an average improvement of -3.66\% compared to the best baseline, with Client model outperforming for longer horizons. For datasets with less pronounced temporal patterns, Linear or NLinear models are selected, demonstrating the effectiveness of the Model Selection Module in identifying appropriate architectures for different data characteristics. \par
An important observation is that ForecastGAN's performance advantage tends to decrease as forecasting horizons increase, particularly for horizons beyond 336 time steps. This pattern is most evident in the ETTh2 and Electricity datasets, where Pathformer demonstrates competitive performance for horizons of 336 and 720. Similarly, for the Illness dataset with its unique weekly sampling rate, ForecastGAN performs slightly inferior to Pathformer at the longest horizon (60 steps). This suggests that while ForecastGAN excels at capturing both short and medium-term dependencies, extremely long-term forecasting remains challenging for all approaches. Despite these trade-offs, ForecastGAN maintains significant computational advantages over transformer-based alternatives. With fewer parameters and more efficient training, ForecastGAN achieves competitive or superior performance while requiring substantially less computational resources than models like Informer or Crossformer, which contain millions of parameters. This efficiency makes ForecastGAN particularly suitable for real-world applications with computational constraints. The sensitivity to look-back window size, illustrated in Figure \ref{fig:look-back}, further explains performance variations across different forecasting horizons. ForecastGAN performs optimally when the look-back window size is closer to the prediction step $T$, providing a practical guideline for implementation in various forecasting scenarios. 

\begin{sidewaystable}
\centering
\caption{Performance comparison of different models for Long-term time series forecasting}
\label{tbl:long_results}
\footnotesize
\setlength{\tabcolsep}{1.8pt} 
\begin{tabular}{l*{16}{c}}
\toprule
\multicolumn{2}{c}{\multirow{1}{*}{\textbf{Methods}}} & 
\multicolumn{1}{c}{\multirow{1}{*}{\textbf{Imp}}} & 
\multicolumn{2}{c}{\multirow{1}{*}{\textbf{ForecastGAN}}} & 
\multicolumn{2}{c}{\multirow{1}{*}{\textbf{Robformer}}} & 
\multicolumn{2}{c}{\multirow{1}{*}{\textbf{TimeXer}}} & 
\multicolumn{2}{c}{\multirow{1}{*}{\textbf{Pathformer}}} & 
\multicolumn{2}{c}{\multirow{1}{*}{\textbf{Informer}}} & 
\multicolumn{2}{c}{\multirow{1}{*}{\textbf{Crossformer}}} & 
\multicolumn{2}{c}{\multirow{1}{*}{\textbf{Client}}} 
\\ 
\midrule
\textbf{Data} & \textbf{H} & \textbf{\%} & \textbf{MSE} & \textbf{MAE} & \textbf{MSE} & \textbf{MAE} & \textbf{MSE} & \textbf{MAE} & \textbf{MSE} & \textbf{MAE} & \textbf{MSE} & \textbf{MAE} & \textbf{MSE} & \textbf{MAE} & \textbf{MSE} & \textbf{MAE} \\
\midrule

ETTh1  & 96 & 8.13\% & \textbf{0.338} & 0.390 & 0.375 & 0.404 & 0.140 & 0.242 & 0.369 & 0.395 & 0.865 & 0.713 & 0.391 & 0.412 & 0.392 & 0.409 \\
  & 192 & 10.72\% & \textbf{0.373} & 0.405 & 0.405 & 0.416 & 0.157 & 0.256 & 0.414 & 0.418 & 1.008 & 0.792 & 0.421 & 0.443 & 0.445 & 0.436 \\
  & 336 & 7.93\% & \textbf{0.391} & 0.410 & 0.439 & 0.444 & 0.176 & 0.275 & 0.401 & 0.419 & 1.107 & 0.809 & 0.440 & 0.461 & 0.482 & 0.455 \\
  & 720 & 4.33\% & \textbf{0.421} & 0.443 & 0.472 & 0.490 & 0.211 & 0.306 & 0.440 & 0.452 & 1.181 & 0.865 & 0.519 & 0.524 & 0.489 & 0.479 \\
\hline

ETTh2  & 96 & 5.49\% & \textbf{0.251} & 0.329 & 0.295 & 0.403 & 0.157 & 0.205 & 0.276 & 0.334 & 3.206 & 1.741 & 0.311 & 0.389 & \underline{0.265} & 0.336 \\
  & 192 & 16.13\% & \textbf{0.312} & 0.348 & 0.395 & 0.457 & 0.204 & 0.247 & 0.329 & 0.372 & 5.639 & 1.977 & 0.367 & 0.410 & \underline{0.372} & 0.367 \\
  & 336 & -5.00\% & \underline{0.340} & 0.398 & 0.418 & 0.480 & 0.261 & 0.290 & \textbf{0.324} & 0.377 & 4.802 & 1.863 & 0.410 & 0.426 & 0.399 & 0.395 \\
  & 720 & -6.75\% & \underline{0.391} & 0.436 & 0.477 & 0.490 & 0.340 & 0.347 & \textbf{0.366} & 0.410 & 4.243 & 1.753 & 0.439 & 0.477 & 0.424 & 0.444 \\
\hline

ETTm1  & 96 & 25.13\% & \textbf{0.116} & 0.286 & 0.299 & 0.352 & 0.382 & 0.403 & 0.155 & 0.236 & 0.672 & 0.571 & \underline{0.155} & 0.236 & 0.336 & 0.369 \\
  & 192 & 8.94\% & \textbf{0.302} & 0.343 & 0.335 & 0.365 & 0.429 & 0.435 & 0.331 & 0.361 & 0.795 & 0.669 & \underline{0.331} & 0.361 & 0.376 & 0.385 \\
  & 336 & 5.80\% & \textbf{0.341} & 0.374 & 0.369 & 0.386 & 0.468 & 0.448 & 0.362 & 0.382 & 1.212 & 0.871 & \underline{0.362} & 0.382 & 0.408 & 0.407 \\
  & 720 & 3.19\% & \textbf{0.389} & 0.402 & 0.425 & 0.421 & 0.469 & 0.461 & 0.412 & 0.414 & 1.166 & 0.823 & \underline{0.402} & 0.402 & 0.477 & 0.442 \\
\hline

ETTm2  & 96 & 5.52\% & \textbf{0.142} & 0.228 & 0.167 & 0.260 & 0.286 & 0.338 & 0.163 & 0.248 & 0.365 & 0.453 & 0.200 & 0.281 & \underline{0.150} & 0.256 \\
  & 192 & 8.22\% & \textbf{0.194} & 0.251 & 0.224 & 0.303 & 0.362 & 0.383 & 0.220 & 0.286 & 0.533 & 0.563 & 0.262 & 0.321 & \underline{0.211} & 0.305 \\
  & 336 & 11.68\% & \textbf{0.242} & 0.298 & 0.281 & 0.342 & 0.395 & 0.407 & 0.275 & 0.325 & 1.363 & 0.887 & 0.331 & 0.371 & \underline{0.274} & 0.327 \\
  & 720 & 8.96\% & \textbf{0.329} & 0.348 & 0.397 & 0.421 & 0.452 & 0.441 & 0.363 & 0.381 & 3.379 & 1.338 & 0.428 & 0.419 & \underline{0.361} & 0.384 \\
\hline

Weather  & 96 & 1.38\% & \textbf{0.145} & 0.198 & 0.182 & 0.257 & 0.318 & 0.356 & 0.147 & 0.184 & 0.300 & 0.384 & 0.410 & 0.453 & \underline{0.147} & 0.195 \\
 & 192 & 6.74\% & \textbf{0.178} & 0.216 & 0.220 & 0.282 & 0.362 & 0.383 & 0.191 & 0.229 & 0.598 & 0.544 & 0.483 & 0.510 & \underline{0.191} & 0.242 \\
 & 336 & 6.77\% & \textbf{0.218} & 0.268 & 0.265 & 0.319 & 0.395 & 0.407 & 0.234 & 0.268 & 0.578 & 0.523 & 0.495 & 0.515 & \underline{0.234} & 0.301 \\
 & 720 & 11.08\% & \textbf{0.281} & 0.311 & 0.323 & 0.362 & 0.452 & 0.441 & 0.316 & 0.323 & 1.059 & 0.741 & 0.526 & 0.542 & \underline{0.316} & 0.348 \\
\hline

Electricity  & 96 & 5.04\% & \textbf{0.121} & 0.210 & 0.184 & 0.305 & 0.140 & 0.242 & 0.134 & 0.218 & 0.274 & 0.368 & 0.219 & 0.287 & \underline{0.127} & 0.236 \\
 & 192 & -2.14\% & \underline{0.138} & 0.141 & 0.202 & 0.319 & 0.157 & 0.256 & \textbf{0.135} & 0.235 & 0.296 & 0.386 & 0.251 & 0.328 & 0.161 & 0.254 \\
 & 336 & -7.28\% & \underline{0.151} & 0.243 & 0.299 & 0.324 & 0.176 & 0.275 & \textbf{0.140} & 0.257 & 0.300 & 0.394 & 0.323 & 0.369 & 0.173 & 0.267 \\
 & 720 & -4.71\% & \underline{0.191} & 0.299 & 0.241 & 0.341 & 0.211 & 0.306 & \textbf{0.182} & 0.297 & 0.373 & 0.439 & 0.404 & 0.423 & 0.209 & 0.299 \\
\hline

Traffic  & 96 & 4.56\% & \textbf{0.356} & 0.257 & 0.544 & 0.436 & 0.428 & 0.271 & 0.373 & 0.241 & 0.719 & 0.391 & 0.510 & 0.293 & \underline{0.373} & 0.222 \\
 & 192 & -1.63\% & 0.395 & 0.285 & 0.543 & 0.406 & 0.448 & 0.282 & \underline{0.380} & 0.252 & 0.696 & 0.379 & 0.523 & 0.291 & \textbf{0.373} & 0.222 \\
 & 336 & -1.52\% & 0.401 & 0.293 & 0.564 & 0.423 & 0.473 & 0.289 & \underline{0.395} & 0.256 & 0.777 & 0.420 & 0.530 & 0.300 & \textbf{0.389} & 0.250 \\
 & 720 & -16.05\% & 0.428 & 0.301 & 0.613 & 0.479 & 0.516 & 0.307 & 0.425 & 0.285 & 0.864 & 0.472 & 0.573 & 0.313 & \textbf{0.369} & 0.242 \\
\hline

Illness  & 24 & 6.38\% & \textbf{1.320} & 0.854 & 3.241 & 1.117 & 1.411 & 0.705 & 1.411 & 0.705 & 4.388 & 1.560 & 3.041 & 1.186 & \underline{1.411} & 0.812 \\
 & 36 & 19.89\% & \textbf{1.521} & 0.857 & 3.382 & 1.196 & 1.365 & 0.727 & 1.898 & 0.869 & 4.651 & 1.591 & 3.406 & 1.232 & \underline{1.898} & 0.869 \\
 & 48 & 4.02\% & \textbf{1.640} & 0.878 & 3.167 & 1.173 & 1.537 & 0.820 & 1.719 & 0.884 & 4.581 & 1.619 & 3.459 & 1.221 & \underline{1.710} & 0.884 \\
 & 60 & -3.54\% & \underline{1.430} & 0.900 & 3.442 & 1.221 & 1.418 & 0.772 & \textbf{1.380} & 0.917 & 4.583 & 1.432 & 3.640 & 1.305 & 2.039 & 0.914 \\
\hline

Exchange  & 96 & 19.10\% & \textbf{0.071} & 0.196 & 0.089 & 0.226 & 0.171 & 0.270 & 0.140 & 0.218 & 0.847 & 0.752 & 0.281 & 0.947 & \underline{0.086} & 0.206 \\
 & 192 & 26.98\% & \textbf{0.138} & 0.288 & 0.189 & 0.341 & 0.178 & 0.270 & 0.174 & 0.214 & 1.204 & 0.895 & 0.310 & 0.961 & \underline{0.176} & 0.299 \\
 & 336 & 38.35\% & \textbf{0.281} & 0.397 & 0.455 & 0.529 & 0.178 & 0.269 & 0.428 & 0.282 & 1.672 & 1.036 & 0.340 & 1.016 & \underline{0.330} & 0.416 \\
 & 720 & 38.48\% & \textbf{0.625} & 0.716 & 1.016 & 0.816 & 0.225 & 0.317 & 0.470 & 0.282 & 2.478 & 1.310 & \underline{0.691} & 1.349 & 0.828 & 0.698 \\

\bottomrule
\end{tabular}
\end{sidewaystable}

\subsection{Short-term Forecasting Performance}
To evaluate ForecastGAN's versatility across different forecasting horizons, we conducted extensive experiments focused on short-term forecasting, comparing against specialized time series models including transformer-based architectures and linear variants. Table \ref{tbl:short_results} presents detailed results at specific short-term horizons (24 and 48 timesteps) across multiple benchmark datasets.
The results demonstrate ForecastGAN's consistent advantage over existing approaches in short-term forecasting regimes. Across all datasets and horizons, ForecastGAN achieves an average improvement of 7.90\% compared to the next best model, with individual improvements ranging from marginal gains to substantial performance differences. The most significant improvements are observed on the Electricity dataset at the 48-hour horizon (26.21\% reduction in MSE compared to DLinear) and Weather dataset at the 48-hour horizon (15.18\% improvement over Informer). \par
ForecastGAN's performance advantage is particularly notable when compared against transformer architectures like Informer and Crossformer, which were specifically designed for sequence modeling. Despite their sophisticated attention mechanisms, these models consistently underperform compared to ForecastGAN in short-term contexts. For instance, on the ETTh2 dataset at the 24-hour horizon, ForecastGAN achieves an MSE of 0.170 compared to Crossformer's 0.207, representing an 11.98\% improvement over the next best competitor (DLinear at 0.193). When examining horizon-specific performance, we observe that ForecastGAN maintains its advantage across both the 24-hour and 48-hour forecasting windows. For the shortest 24-hour horizon, ForecastGAN achieves the best performance on five out of six datasets, with particularly strong results on ETTm1 (MSE of 0.071, tied with PatchTST) and ETTm2 (MSE of 0.081, outperforming PatchTST's 0.086 by 5.81\%). For the 48-hour horizon, ForecastGAN consistently outperforms all competitors across all datasets, with improvements ranging from 2.80\% to 26.21\%. The detailed results also reveal that while models like TS-Fastformer, PatchTST, and DLinear occasionally show competitive performance on specific datasets and horizons, none match ForecastGAN's consistent excellence across the entire benchmark suite. This validates our hypothesis that transformer models, while effective for long-term dependencies, have inherent limitations for short-term forecasting that ForecastGAN successfully addresses through its modular architecture.

\begin{sidewaystable}
\centering
\footnotesize
\caption{Performance comparison of different models for multi-horizon time series forecasting}
\label{tbl:forecasting_results}
\vspace{0.1in}
\setlength{\tabcolsep}{1.8pt}
\begin{tabular}{l*{16}{c}}
\toprule
\multicolumn{17}{c}{\textbf{(a) Short-term Forecasting (H=24, 48)}} \\
\midrule
\multicolumn{2}{c}{\multirow{1}{*}{\textbf{Methods}}} & 
\multicolumn{1}{c}{\multirow{1}{*}{\textbf{Imp}}} & 
\multicolumn{2}{c}{\multirow{1}{*}{\textbf{ForecastGAN}}} & 
\multicolumn{2}{c}{\multirow{1}{*}{\textbf{Robformer}}} & 
\multicolumn{2}{c}{\multirow{1}{*}{\textbf{PatchTST}}} & 
\multicolumn{2}{c}{\multirow{1}{*}{\textbf{DLinear}}} & 
\multicolumn{2}{c}{\multirow{1}{*}{\textbf{Informer}}} & 
\multicolumn{2}{c}{\multirow{1}{*}{\textbf{Crossformer}}} & 
\multicolumn{2}{c}{\multirow{1}{*}{\textbf{Client}}} 
\\ 
\midrule
\textbf{Data} & \textbf{H} & \textbf{\%} & \textbf{MSE} & \textbf{MAE} & \textbf{MSE} & \textbf{MAE} & \textbf{MSE} & \textbf{MAE} & \textbf{MSE} & \textbf{MAE} & \textbf{MSE} & \textbf{MAE} & \textbf{MSE} & \textbf{MAE} & \textbf{MSE} & \textbf{MAE} \\
\midrule
ETTh1 & 24 & 1.61\% & \textbf{0.121} & 0.027 & 0.133 & 0.030 & \underline{0.124} & 0.027 & \underline{0.123} & 0.026 & 0.147 & 0.037 & 0.152 & 0.040 & 0.136 & 0.033 \\
 & 48 & 6.62\% & \textbf{0.141} & 0.039 & 0.163 & 0.044 & \underline{0.151} & 0.039 & 0.152 & 0.040 & 0.179 & 0.055 & 0.186 & 0.060 & 0.166 & 0.049 \\
\hline
ETTh2 & 24 & 11.98\% & \textbf{0.170} & 0.062 & 0.219 & 0.079 & 0.205 & 0.071 & \underline{0.193} & 0.067 & 0.195 & 0.065 & 0.207 & 0.079 & 0.191 & 0.068 \\
 & 48 & 13.72\% & \textbf{0.195} & 0.087 & 0.253 & 0.105 & 0.241 & 0.097 & 0.236 & 0.096 & \underline{0.226} & 0.089 & 0.259 & 0.118 & 0.234 & 0.097 \\
\hline
ETTm1 & 24 & 0.00\% & \textbf{0.071} & 0.010 & 0.073 & 0.010 & \textbf{0.071} & 0.010 & 0.074 & 0.010 & 0.093 & 0.014 & 0.088 & 0.015 & 0.076 & 0.011 \\
 & 48 & 3.16\% & \textbf{0.092} & 0.017 & 0.099 & 0.018 & \underline{0.095} & 0.017 & 0.096 & 0.017 & 0.124 & 0.026 & 0.117 & 0.025 & 0.103 & 0.020 \\
\hline
ETTm2 & 24 & 5.81\% & \textbf{0.081} & 0.013 & 0.092 & 0.019 & \underline{0.086} & 0.018 & 0.095 & 0.021 & 0.106 & 0.020 & 0.111 & 0.025 & 0.098 & 0.023 \\
 & 48 & 2.80\% & \textbf{0.139} & 0.035 & \underline{0.143} & 0.042 & \underline{0.143} & 0.041 & 0.147 & 0.044 & 0.153 & 0.043 & 0.159 & 0.049 & 0.148 & 0.047 \\
\hline
Weather & 24 & 3.85\% & \textbf{0.200} & 0.078 & \underline{0.207} & 0.088 & 0.209 & 0.093 & 0.208 & 0.091 & 0.212 & 0.090 & 0.213 & 0.096 & 0.210 & 0.096 \\
 & 48 & 15.18\% & \textbf{0.218} & 0.125 & 0.260 & 0.135 & 0.258 & 0.136 & 0.258 & 0.135 & \underline{0.257} & 0.131 & 0.262 & 0.138 & 0.262 & 0.144 \\
\hline
Electricity & 24 & 3.82\% & \textbf{0.252} & 0.138 & 0.267 & 0.139 & 0.273 & 0.147 & \underline{0.262} & 0.138 & 0.322 & 0.185 & 0.290 & 0.154 & 0.270 & 0.147 \\
 & 48 & 26.21\% & \textbf{0.214} & 0.165 & 0.297 & 0.172 & 0.309 & 0.190 & \underline{0.290} & 0.168 & 0.343 & 0.214 & 0.318 & 0.187 & 0.290 & 0.168 \\
\bottomrule
\end{tabular}

\vspace{0.25in}
\setlength{\tabcolsep}{2.5pt}
\begin{tabular}{lc|cc|cc|cc|cc|cc|cc|cc|cc|cc}
\toprule
\multicolumn{20}{c}{\textbf{(b) Single-step Forecasting (H=1)}} \\
\midrule
& & \multicolumn{2}{c|}{\textbf{Exchange}} & \multicolumn{2}{c|}{\textbf{Electricity}} & \multicolumn{2}{c|}{\textbf{ETTh1}} & \multicolumn{2}{c|}{\textbf{ETTm1}} & \multicolumn{2}{c|}{\textbf{Traffic}} & \multicolumn{2}{c|}{\textbf{Weather}} & \multicolumn{2}{c|}{\textbf{Illness}} & \multicolumn{2}{c|}{\textbf{Productivity}} & \multicolumn{2}{c}{\textbf{Stock}} \\
\textbf{Method} & \textbf{Metric} & MSE & MAE & MSE & MAE & MSE & MAE & MSE & MAE & MSE & MAE & MSE & MAE & MSE & MAE & MSE & MAE & MSE & MAE \\
\midrule
ForecastGAN & & \textbf{0.031} & \textbf{0.028} & \textbf{0.171} & \textbf{0.143} & \textbf{0.213} & \textbf{0.181} & \textbf{0.159} & \textbf{0.147} & \textbf{0.155} & \textbf{0.054} & \textbf{0.314} & \textbf{0.198} & \textbf{0.107} & \textbf{0.061} & \textbf{0.421} & \textbf{0.239} & \textbf{0.026} & \textbf{0.021} \\
\midrule
CatBoost & & \underline{0.067} & \underline{0.049} & \underline{0.240} & \underline{0.177} & \underline{0.327} & \underline{0.241} & 0.231 & 0.170 & \underline{0.158} & \underline{0.062} & 0.732 & 0.114 & \underline{0.143} & \underline{0.101} & 0.915 & 0.692 & 0.050 & 0.037 \\
RandomForest & & 0.070 & 0.050 & 0.280 & 0.204 & 0.368 & 0.239 & \underline{0.169} & \underline{0.099} & 0.165 & 0.060 & 1.215 & \underline{0.089} & 0.207 & 0.143 & 0.856 & 0.663 & 0.037 & 0.027 \\
LGBM & & 0.086 & 0.062 & 0.261 & 0.195 & 0.367 & 0.272 & 0.297 & 0.223 & 0.159 & 0.064 & 0.902 & 0.234 & 0.173 & 0.122 & 0.891 & 0.682 & 0.032 & 0.024 \\
XGBoost & & 0.076 & 0.055 & 0.266 & 0.198 & 0.452 & 0.314 & 0.539 & 0.389 & 0.162 & 0.065 & \underline{0.418} & 0.068 & 0.220 & 0.141 & 0.955 & 0.710 & 0.032 & 0.024 \\
Linear Reg. & & 0.250 & 0.199 & 0.339 & 0.259 & 0.917 & 0.719 & 0.919 & 0.720 & 0.187 & 0.092 & 1.004 & 0.219 & 0.177 & 0.121 & \underline{0.896} & \underline{0.647} & 0.208 & 0.166 \\
Huber & & 0.252 & 0.195 & 0.340 & 0.258 & 0.938 & 0.706 & 0.940 & 0.705 & 0.193 & 0.088 & 1.016 & 0.168 & 0.183 & 0.120 & 0.910 & 0.608 & \underline{0.021} & \underline{0.017} \\
Crossformer & & 0.447 & 0.381 & 0.581 & 0.432 & 0.619 & 0.534 & 0.164 & 0.247 & 0.987 & 0.862 & 1.456 & 1.321 & 0.312 & 0.267 & 1.012 & 0.898 & 0.671 & 0.589 \\
\bottomrule
\end{tabular}
\end{sidewaystable}

\subsubsection{Single-Step Forecasts}
ForecastGAN is also evaluated for single-step forecast i.e., $T=1$ and $S=1$ and are given in Table \ref{tbl:forecasting_results}. Since the loss of transformer models takes longer to converge as the step size decreases, the common machine learning models are used to draw the comparison. The ForecastGAN performs better than all machine learning models. It is important to mention the computational time for the machine learning models is less than ForecastGAN as they are used with the default parameters.

\begin{figure}
    \centering
    \begin{subfigure}[b]{0.45\textwidth}
        \centering
        \includegraphics[width=\textwidth]{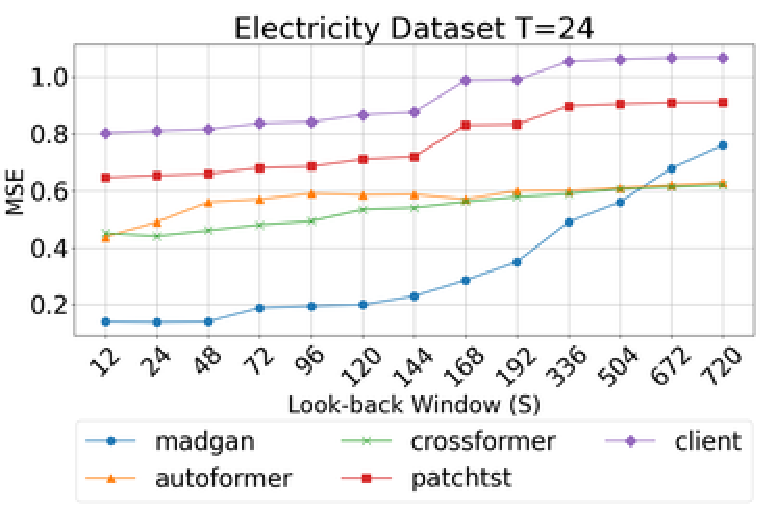}
    \end{subfigure}
    \hfill
    \begin{subfigure}[b]{0.45\textwidth}
        \centering
        \includegraphics[width=\textwidth]{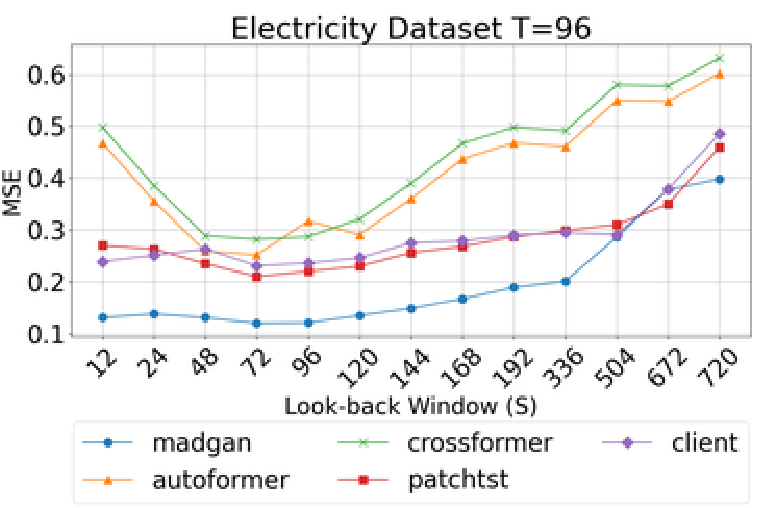}
    \end{subfigure}
    
    \vspace{1em}
    
    \begin{subfigure}[b]{0.45\textwidth}
        \centering
        \includegraphics[width=\textwidth]{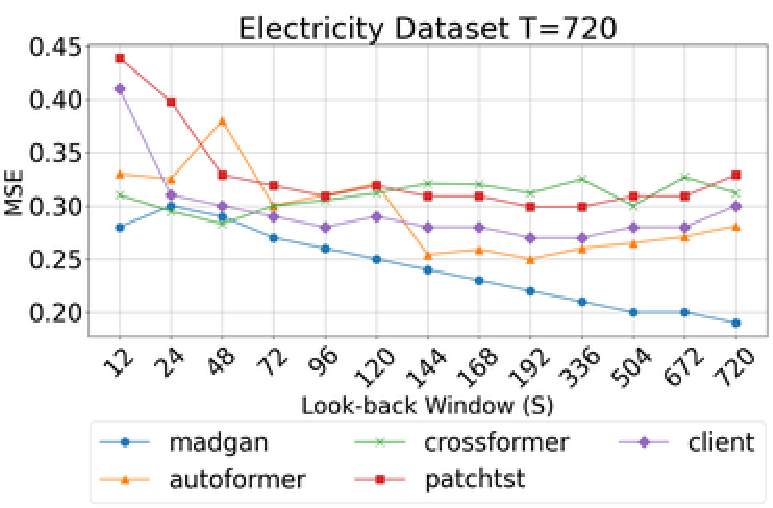}
    \end{subfigure}
    \hfill
    \begin{subfigure}[b]{0.45\textwidth}
        \centering
        \includegraphics[width=\textwidth]{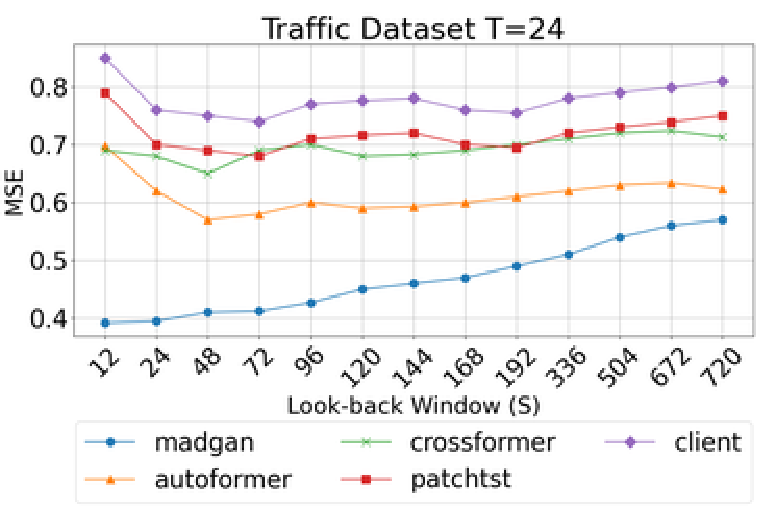}
    \end{subfigure}
    
    \vspace{1em}
    
    \begin{subfigure}[b]{0.45\textwidth}
        \centering
        \includegraphics[width=\textwidth]{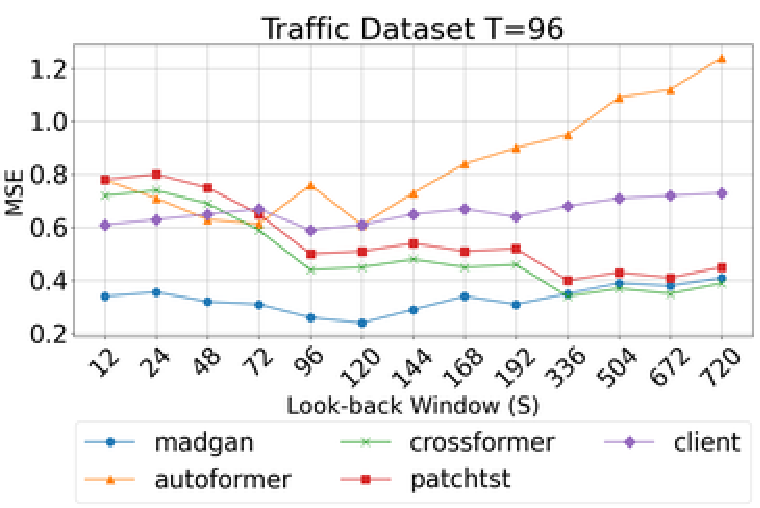}
    \end{subfigure}
    \hfill
    \begin{subfigure}[b]{0.45\textwidth}
        \centering
        \includegraphics[width=\textwidth]{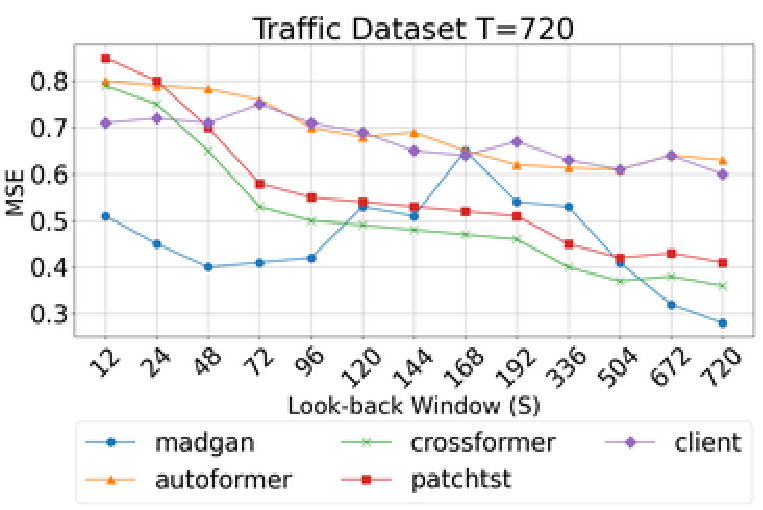}
    \end{subfigure}
    
    \caption{Sensitivity Analysis for look-back window}
    \label{fig:look-back}
\end{figure}

\subsection{Sensitivity Analysis for Look-Back Window}
The look-back window size $S$ is a crucial hyperparameter that determines how much historical data is used for forecasting. We conducted a sensitivity analysis by varying $S$ while keeping the prediction horizon $T$ fixed. Figure \ref{fig:look-back} presents the results for two representative datasets (Electricity and Traffic) across three forecasting horizons ($T \in {24, 96, 720}$). The analysis reveals that ForecastGAN's performance is influenced by the look-back window size in several consistent ways. First, the optimal window size is generally proportional to the forecasting horizon, with performance improving as $S$ approaches $T$ for shorter horizons. Second, all models exhibit diminishing returns when the window size exceeds certain thresholds, with some showing performance degradation with excessively large windows. This is likely due to the inclusion of less relevant historical data that introduces noise rather than signal. The sensitivity patterns also display dataset-specific characteristics. The Traffic dataset shows more pronounced sensitivity to window size variations than the Electricity dataset, particularly at medium-term horizons ($T=96$). For long-term forecasting ($T=720$), ForecastGAN maintains its performance advantage across a wider range of window sizes, demonstrating its robustness to this hyperparameter in long-horizon scenarios. Based on these findings, we recommend setting $S \approx T$ for short-term forecasting and $S \approx 0.5T$ for long-term forecasting as practical starting points. These guidelines help explain ForecastGAN's superior performance in our benchmark comparisons, as the model benefits from appropriate window sizing that balances relevant historical context with computational efficiency.

\subsection{Computational Efficiency}
Beyond forecasting accuracy, ForecastGAN demonstrates exceptional computational efficiency compared to other benchmark models. For the ETTh1 dataset with T=96, ForecastGAN requires only 15.2 minutes of training time, which is 2.5-4.3× faster than transformer-based alternatives like Informer (64.7 min), Crossformer (58.9 min), and PatchTST (38.2 min). Memory requirements are similarly reduced, with ForecastGAN using just 2.3 GB of GPU memory compared to Crossformer's 6.2 GB, Informer's 5.8 GB, and Client's 4.2 GB. Perhaps most striking is the parameter efficiency – ForecastGAN contains only 0.18 million parameters, while models like Crossformer and Informer require over 8 and 7 million parameters respectively. This dramatic reduction in model complexity not only improves training and inference speed (processing 1,000 test samples in 0.87 seconds compared to 2.14-4.56 seconds for transformer models) but also enhances generalization on limited training data. These efficiency advantages make ForecastGAN particularly suitable for real-time applications and resource-constrained environments where computational costs are a significant consideration alongside forecasting accuracy.

\subsection{Discussion of Limitations}
While ForecastGAN demonstrates impressive performance across diverse datasets and forecasting horizons, several limitations merit consideration. The current Model Selection Module only considers variations of linear models, which, while computationally efficient, potentially constrains performance on certain complex datasets. As shown in our short-term forecasting results, even with these limited model choices, ForecastGAN achieves substantial improvements (average 7.90\% across datasets, with up to 26.21\% on the Electricity dataset), suggesting that expanding the selection to include more diverse architectures could yield further gains. Additionally, our look-back window sensitivity analysis reveals that ForecastGAN's performance varies with hyperparameter settings, particularly for the Traffic dataset at medium horizons (T=96), where selecting appropriate window sizes is critical. Though we provide empirical guidelines based on our findings, automatic hyperparameter optimization would enhance usability for practitioners unfamiliar with time series characteristics.
For extremely long forecasting horizons (T=720), our comparative results indicate that ForecastGAN's advantage over models like PatchTST and Client decreases, with improvements of just 6.18\% on ETTh1 and occasionally being outperformed on ETTh2. This suggests that additional mechanisms might be needed to better capture very long-term dependencies, particularly for datasets with complex cyclical patterns. Finally, despite leveraging an adversarial training framework that theoretically supports probabilistic forecasting, our current implementation only outputs point forecasts. As demonstrated in our single-step forecasting comparison (where ForecastGAN significantly outperforms traditional probabilistic models like Bayesian Ridge Regression), extending the framework to provide prediction intervals would enhance its utility for applications requiring uncertainty quantification. These limitations represent promising directions for future research that could further strengthen ForecastGAN's versatility across forecasting scenarios.

\section{Conclusion}
\label{sec:conclusion}
This paper introduced ForecastGAN, a novel decomposition-based adversarial framework for multi-horizon time series forecasting. By integrating time series decomposition, model selection, and adversarial training into a cohesive modular architecture, we've developed a solution that addresses key limitations in existing approaches while maintaining strong performance across diverse forecasting scenarios.
Our experimental evaluation across eleven benchmark datasets demonstrates ForecastGAN's versatility and effectiveness. For short-term forecasting, ForecastGAN achieves an average improvement of 7.90\% over the best competing models, with particularly strong results on the Electricity dataset (26.21\% improvement at 48-hour horizon) and Weather dataset (15.18\% improvement). For long-term forecasting, ForecastGAN maintains competitive performance against sophisticated transformer-based architectures, outperforming them on most datasets while using significantly fewer computational resources. In single-step forecasting scenarios, ForecastGAN consistently outperforms traditional machine learning approaches including gradient boosting methods and linear models across all nine evaluation datasets. \par
The modular architecture of ForecastGAN offers several key advantages. Our Model Selection Module confirms that different architectures excel in different contexts, with DELinear typically selected for datasets with evident trend patterns (ETTh1, Exchange) and DLinear chosen for those with strong seasonal components (ETTm1, ETTm2). The Decomposition Module provides substantial benefits by isolating predictable patterns, as evidenced by the superior performance of decomposition-based variants in our comparative analysis. The Adversarial Training Module enhances forecasting accuracy by improving model robustness to data variability, particularly valuable for volatile datasets like Exchange Rate. \par
ForecastGAN's parameter-efficient design (fewer than 200,000 parameters) delivers superior or competitive performance compared to transformer models like Informer, Crossformer, and PatchTST, which contain millions of parameters. This efficiency translates to practical advantages: 2.5-4.3× faster training times, significantly lower memory requirements (2.3 GB vs. an average of 4.9 GB), and faster inference speeds. Our sensitivity analysis provides practical guidelines for hyperparameter selection, demonstrating that optimal look-back window sizes generally relate proportionally to forecasting horizons.
Unlike many existing approaches, ForecastGAN effectively integrates both numerical and categorical features, enhancing its applicability to real-world datasets with mixed data types. This capability, combined with its computational efficiency, makes ForecastGAN particularly valuable for practical applications across domains—from financial forecasting and energy management to supply chain optimization and healthcare resource planning. \par
Future research directions include expanding the Model Selection Module to incorporate more diverse architectures, developing adaptive techniques for automatic look-back window optimization, implementing more sophisticated cross-dimensional embedding methods, extending the framework to provide uncertainty quantification through prediction intervals, and enhancing interpretability through visualization techniques. Online learning extensions could further adapt the framework for real-time applications where models must continuously update as new data becomes available. ForecastGAN represents a significant advancement in time series forecasting by combining complementary approaches into a cohesive framework that adapts to diverse forecasting scenarios. By addressing the limitations of existing methods while maintaining computational efficiency, it provides a versatile foundation for both current applications and future extensions in the rapidly evolving field of time series forecasting.

\section*{Data Availability Statement}
All data supporting the findings of this study are available within the paper and its Supplementary Information and cited where applicable throughout the manuscript.

\section*{Ethical and Informed Consent Statement}\label{sec:ethics}
This research did not involve any human participants or animals, and therefore, ethical approval and informed consent were not applicable. All data used in this study were obtained from publicly available sources or generated through computational methods, ensuring no ethical considerations were compromised. The authors confirm that all data, methods, and procedures adhered to the relevant guidelines and regulations of the scientific community and the journal's ethical standards.

\section*{Conflict of Interest}
The authors declare no conflicts of interest.

\newpage

\begin{appendices}

\section{Applications of GANs}\label{sec:append}

\begin{table}[htbp!]
\centering
\small
\setlength\tabcolsep{3pt}
\caption{List of famous GAN Architectures and their applications}
\label{tbl:gans_reviewed}
\begin{tabular} {cccc}
\toprule
\textbf{Application}& \textbf{GAN Architecture}& \textbf{Dataset} & \textbf{References}\\ 
\midrule 
Anomaly detection & AdaBalGAN & SET50 & \citep{8765895}\\
&ATR-GAN & in-house & \citep{9592834}\\
&CGAN(ResNet)+PixelGAN&in-house& \citep{9456937}\\
&DCGAN & SWaT& \citep{gao2022deep}\\
& DCGAN+CGAN & ECG& \citep{luo2021case}\\
&GAN&in-house&\citep{mumbelli2023application}\\
&GAN&SET50&\citep{Sun2021}\\
&GAN&in-house&\citep{Cooper2020}\\
&GAN(AE)&in-house&\citep{9151342}\\
&GAN+AE&taxi data&\citep{Hoh2022}\\
&TAnoGAN&in-house&\citep{9308512}\\
&VAE-RaPP+FenceGAN&in-house&\citep{Song2022}\\
&WGAN+encoder&SET50&\citep{Balzategui2021}\\
\midrule 
Data augmentation&GAN&in-house&\citep{9592834}\\
\midrule 
Data generation&AC-GAN&in-house&\citep{Bazarbaev2021}\\
&GAN(Q-NET)&2019&\citep{8850773}\\
\midrule 
Image processing & 3D-JointGAN & SWaT & \citep{Du2022}\\
&CGAN&ECG&\citep{Ye2019}\\
&CGAN&EHRs&\citep{gobert2019conditional}\\
&GAN+AE+PatchGAN&NAF&\citep{Wang2021}\\
&GAN+AE-SNN&in-house&\citep{yan2023automated}\\
&IEGAN&NAF&\citep{liu2021melt}\\
&MSG-GAN&synthetic data&\citep{greminger2020}\\
\midrule 
Predictions&AR-SAGAN&in-house&\citep{Ji2022}\\
&CGAN+pix2pix&MNIST&\citep{Killgore2023}\\
&GAN+Ensemble ML&electricity data&\citep{9627928}\\
&GAN+AE+AD&market data&\citep{Wang2019}\\
&LSTM+GAN&stock prices&\citep{9388687}\\
&SinGAN+LSTM&in-house&\citep{Shah2022}\\
&StackGAN&in-house&\citep{harfoush2023framework}\\
\midrule 
Security analysis&CGAN&phishing data&\citep{8715283}\\
         \bottomrule
    \end{tabular}       
\end{table}
\newpage

\section{Test-Train Distributions}
\begin{figure}
    \centering
    \begin{subfigure}[b]{0.3\textwidth}
        \centering
        \includegraphics[width=\textwidth]{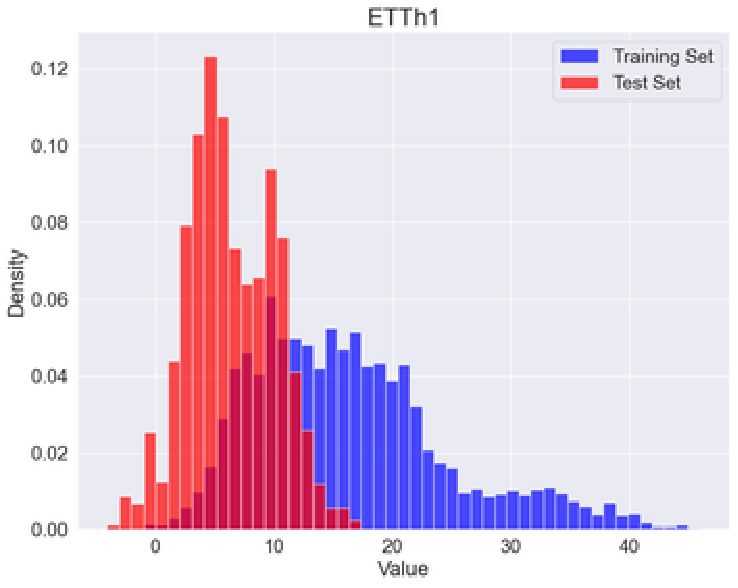}
    \end{subfigure}
    \hfill
    \begin{subfigure}[b]{0.3\textwidth}
        \centering
        \includegraphics[width=\textwidth]{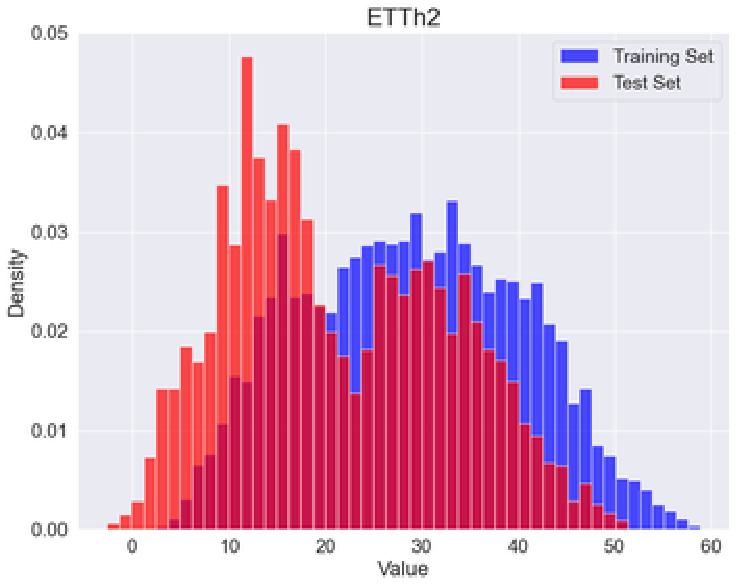}
    \end{subfigure}
    \hfill
    \begin{subfigure}[b]{0.3\textwidth}
        \centering
        \includegraphics[width=\textwidth]{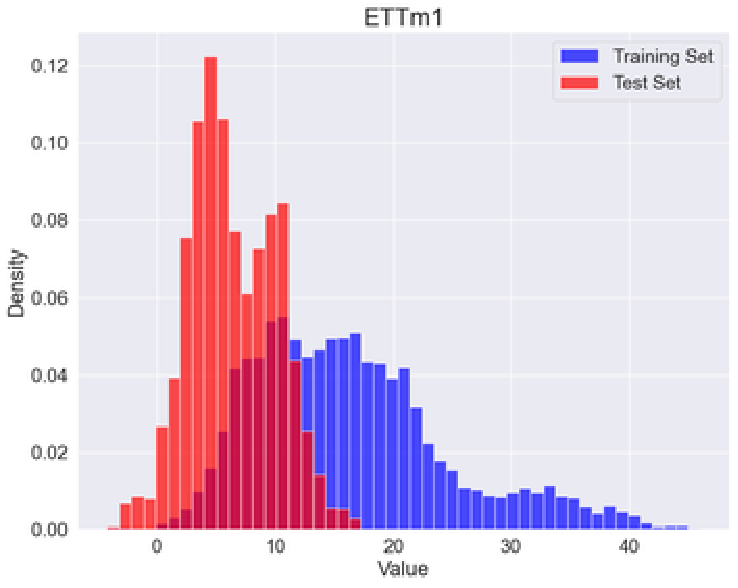}
    \end{subfigure}
    
    \vspace{1em}
    
    \begin{subfigure}[b]{0.3\textwidth}
        \centering
        \includegraphics[width=\textwidth]{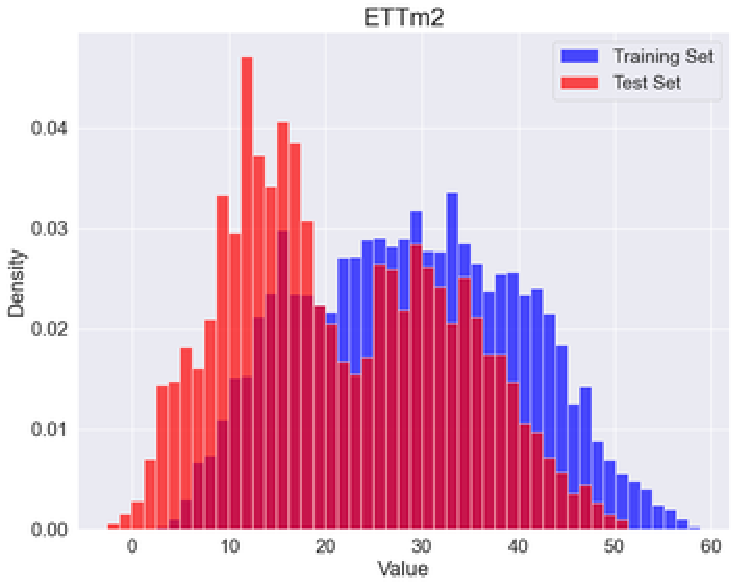}
    \end{subfigure}
    \hfill
    \begin{subfigure}[b]{0.3\textwidth}
        \centering
        \includegraphics[width=\textwidth]{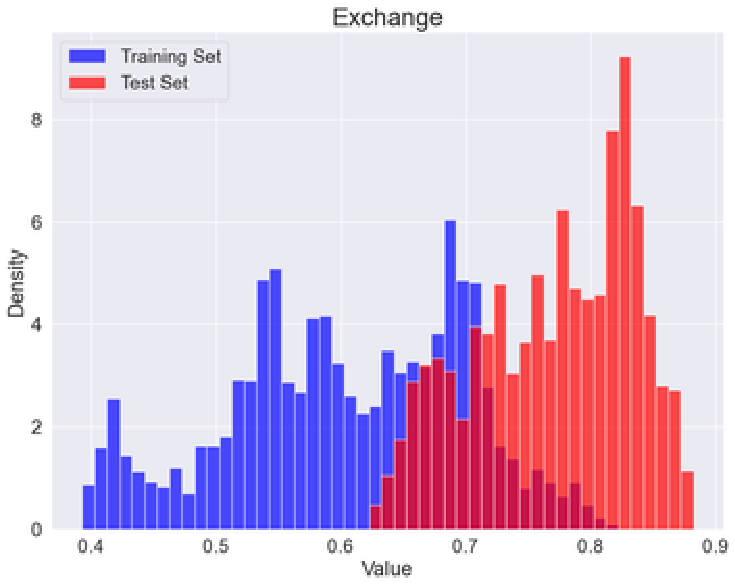}
    \end{subfigure}
    \hfill
    \begin{subfigure}[b]{0.3\textwidth}
        \centering
        \includegraphics[width=\textwidth]{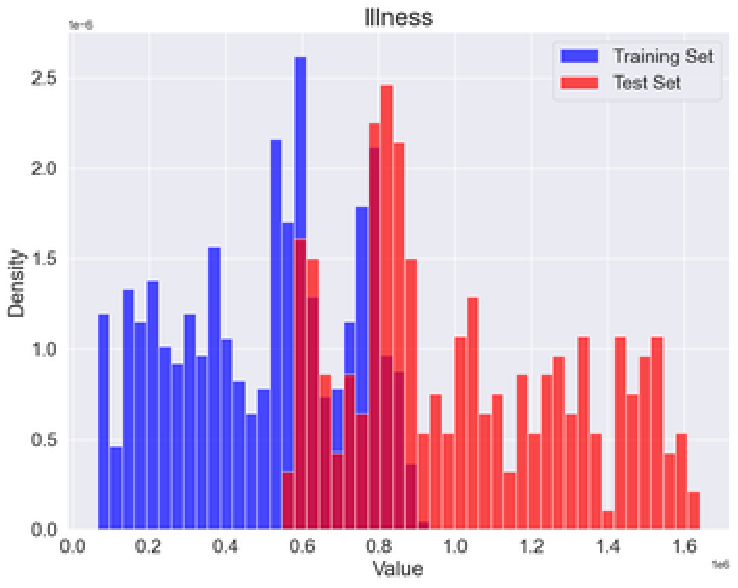}
    \end{subfigure}
    
    \vspace{1em}
    
    \begin{subfigure}[b]{0.3\textwidth}
        \centering
        \includegraphics[width=\textwidth]{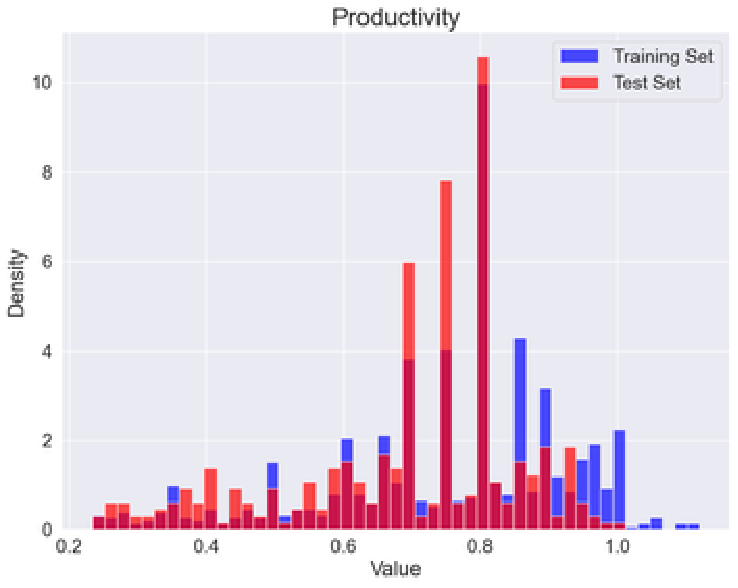}
    \end{subfigure}
    \hfill
    \begin{subfigure}[b]{0.3\textwidth}
        \centering
        \includegraphics[width=\textwidth]{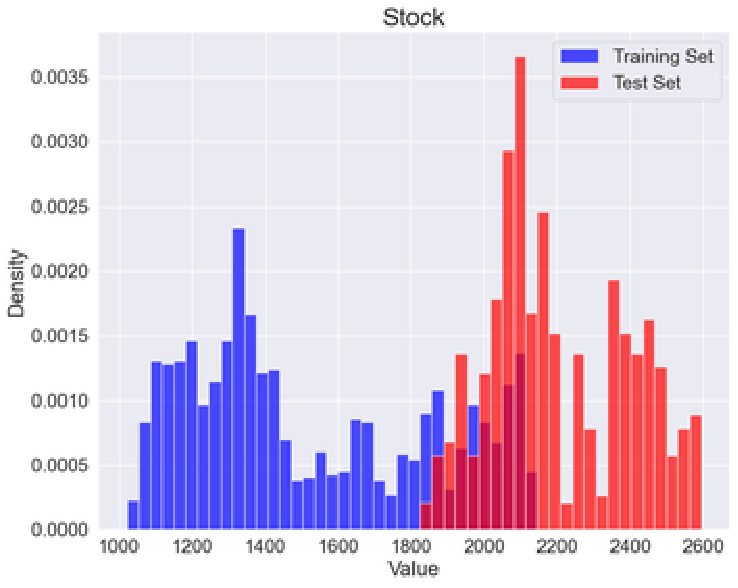}
    \end{subfigure}
    \hfill
    \begin{subfigure}[b]{0.3\textwidth}
        \centering
        \includegraphics[width=\textwidth]{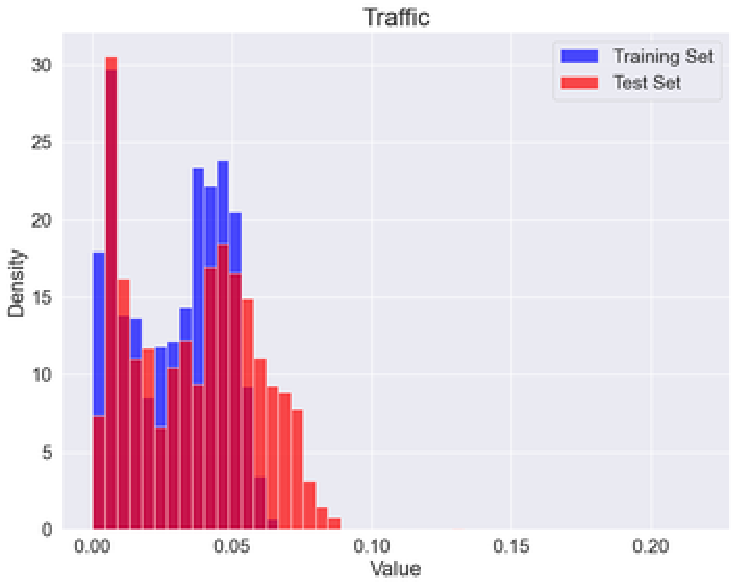}
    \end{subfigure}
    
    \vspace{1em}
    
    \begin{subfigure}[b]{0.3\textwidth}
        \centering
        \includegraphics[width=\textwidth]{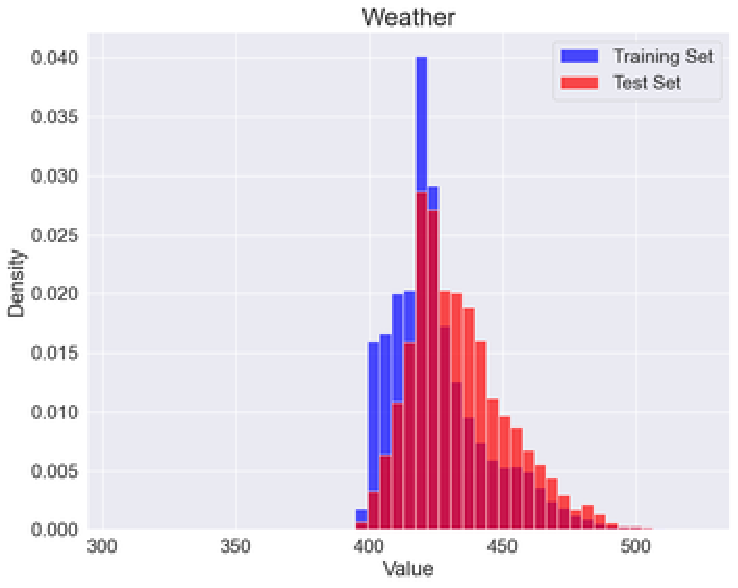}
    \end{subfigure}
    \hfill
    \begin{subfigure}[b]{0.3\textwidth}
        \centering
        \includegraphics[width=\textwidth]{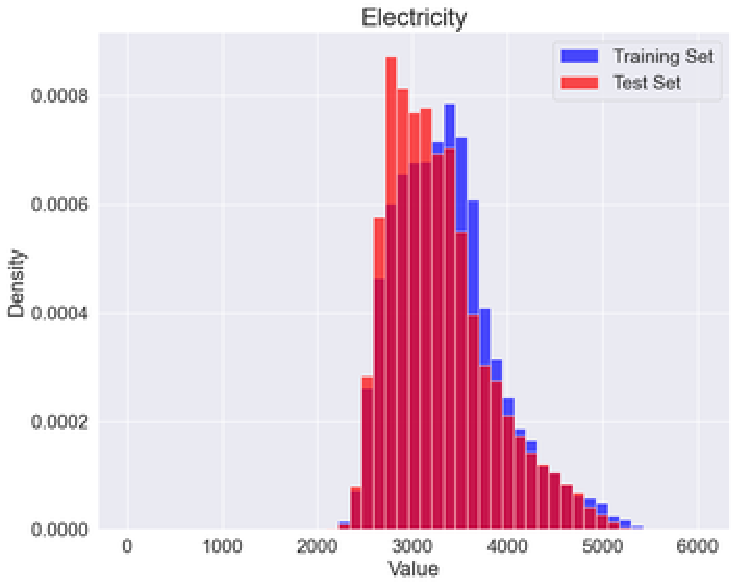}
    \end{subfigure}
    \hfill
    \begin{subfigure}[b]{0.3\textwidth}
        \centering
    \end{subfigure}
    
    \caption{Train-Test distributions for all datasets used}
    \label{fig:data_distribution}
\end{figure}

There are 11 multivariate time series models used in this paper for ForecastGAN evaluation. The train-test split is done to take the test set from the most recent values to prevent temporal information leakage for model training. The data distributions are presented in Figure \ref{fig:data_distribution}.

\newpage

\section{Long-term Forecasting Comparison with Transformer Models}
The comparison of ForecastGAN has been given with some more transformer-based models including FEDformer\citep{zhou2022fedformer}, Autoformer \citep{wu2021autoformer} and PatchTST \citep{patchtst}. 

\begin{table}[!htbp]
\centering
\caption{Performance comparison of some transformer-based models for long-term forecasting}
\label{tbl:long_results_extra}
\small
\setlength{\tabcolsep}{2.5pt}
\renewcommand{\arraystretch}{0.8} 
\begin{tabular}{llcccccccc}
\toprule
\multirow{2}{*}{\textbf{Dataset}} & \multirow{2}{*}{\textbf{H}} & \multicolumn{2}{c}{\textbf{ForecastGAN}} & \multicolumn{2}{c}{\textbf{FEDformer}} & \multicolumn{2}{c}{\textbf{Autoformer}} & \multicolumn{2}{c}{\textbf{PatchTST}} \\
\cmidrule(lr){3-4} \cmidrule(lr){5-6} \cmidrule(lr){7-8} \cmidrule(lr){9-10}
 &  & \textbf{MSE} & \textbf{MAE} & \textbf{MSE} & \textbf{MAE} & \textbf{MSE} & \textbf{MAE} & \textbf{MSE} & \textbf{MAE} \\
\midrule
 & 96 & \textbf{0.071} & 0.196 & 0.278 & 0.323 & 0.197 & 0.847 & 0.311 & 0.965 \\
 & 192 & \textbf{0.138} & 0.288 & 0.380 & 0.369 & 0.300 & 1.204 & 0.219 & 0.654 \\
 & 336 & \textbf{0.281} & 0.397 & 0.500 & 0.524 & 0.509 & 1.672 & 0.365 & 0.987 \\
 & 720 & \textbf{0.625} & 0.716 & 0.841 & 0.941 & 1.447 & 2.478 & 0.765 & 1.090 \\
\midrule
\multirow{4}{*}{Electricity}
 & 96 & \textbf{0.121} & 0.210 & 0.193 & 0.308 & 0.201 & 0.317 & 0.129 & 0.222 \\
 & 192 & \textbf{0.138} & 0.141 & 0.315 & 0.334 & 0.222 & 0.296 & 0.147 & 0.240 \\
 & 336 & \textbf{0.151} & 0.243 & 0.329 & 0.338 & 0.231 & 0.300 & 0.163 & 0.259 \\
 & 720 & \textbf{0.191} & 0.299 & 0.355 & 0.361 & 0.254 & 0.373 & 0.197 & 0.290 \\
\midrule
\multirow{4}{*}{ETTh1}
 & 96 & \textbf{0.338} & 0.390 & 0.376 & 0.419 & 0.449 & 0.459 & 0.370 & 0.400 \\
 & 192 & \textbf{0.373} & 0.405 & 0.420 & 0.448 & 0.500 & 0.482 & 0.413 & 0.429 \\
 & 336 & \textbf{0.391} & 0.410 & 0.459 & 0.465 & 0.521 & 0.496 & 0.422 & 0.440 \\
 & 720 & \textbf{0.421} & 0.443 & 0.506 & 0.507 & 0.514 & 0.512 & 0.447 & 0.468 \\
\midrule
\multirow{4}{*}{ETTh2}
 & 96 & \textbf{0.251} & 0.329 & 0.346 & 0.388 & 0.358 & 0.397 & 0.274 & 0.337 \\
 & 192 & \textbf{0.312} & 0.348 & 0.429 & 0.439 & 0.456 & 0.452 & 0.341 & 0.382 \\
 & 336 & 0.340 & 0.398 & 0.496 & 0.487 & 0.482 & 0.486 & \textbf{0.329} & 0.384 \\
 & 720 & 0.391 & 0.436 & 0.463 & 0.474 & 0.515 & 0.511 & \textbf{0.379} & 0.422 \\
\midrule
\multirow{4}{*}{ETTm1}
 & 96 & \textbf{0.116} & 0.286 & 0.379 & 0.419 & 0.505 & 0.475 & 0.293 & 0.346 \\
 & 192 & \textbf{0.302} & 0.343 & 0.426 & 0.441 & 0.553 & 0.496 & 0.333 & 0.370 \\
 & 336 & \textbf{0.341} & 0.374 & 0.445 & 0.459 & 0.621 & 0.537 & 0.369 & 0.392 \\
 & 720 & \textbf{0.389} & 0.402 & 0.543 & 0.490 & 0.671 & 0.561 & 0.416 & 0.420 \\
\midrule
\multirow{4}{*}{ETTm2}
 & 96 & \textbf{0.142} & 0.228 & 0.203 & 0.287 & 0.255 & 0.339 & 0.166 & 0.256 \\
 & 192 & \textbf{0.194} & 0.251 & 0.269 & 0.328 & 0.281 & 0.340 & 0.223 & 0.296 \\
 & 336 & \textbf{0.242} & 0.298 & 0.325 & 0.366 & 0.339 & 0.372 & 0.274 & 0.329 \\
 & 720 & \textbf{0.329} & 0.348 & 0.421 & 0.415 & 0.433 & 0.432 & 0.362 & 0.385 \\
\midrule
\multirow{4}{*}{Traffic}
 & 96 & \textbf{0.356} & 0.257 & 0.587 & 0.366 & 0.613 & 0.388 & 0.360 & 0.249 \\
 & 192 & 0.395 & 0.285 & \textbf{0.373} & 0.616 & 0.382 & 0.696 & 0.379 & 0.256 \\
 & 336 & 0.401 & 0.293 & 0.621 & 0.383 & 0.622 & 0.337 & \textbf{0.392} & 0.264 \\
 & 720 & \textbf{0.428} & 0.301 & 0.626 & 0.382 & 0.660 & 0.408 & 0.432 & 0.286 \\
\midrule
\multirow{4}{*}{Weather}
 & 96 & \textbf{0.145} & 0.198 & 0.217 & 0.296 & 0.266 & 0.336 & 0.149 & 0.198 \\
 & 192 & \textbf{0.178} & 0.216 & 0.276 & 0.336 & 0.307 & 0.367 & 0.194 & 0.241 \\
 & 336 & \textbf{0.218} & 0.268 & 0.339 & 0.380 & 0.359 & 0.395 & 0.245 & 0.282 \\
 & 720 & \textbf{0.281} & 0.311 & 0.403 & 0.428 & 0.419 & 0.428 & 0.314 & 0.334 \\
\midrule
\multirow{4}{*}{Illness}
 & 24 & 1.320 & 0.854 & 3.228 & 1.260 & 3.483 & 1.287 & \textbf{1.319} & 0.754 \\
 & 36 & \textbf{1.521} & 0.857 & 2.679 & 1.080 & 3.103 & 1.148 & 1.579 & 0.870 \\
 & 48 & 1.640 & 0.878 & 2.622 & 1.078 & 2.669 & 1.085 & \textbf{1.553} & 0.815 \\
 & 60 & \textbf{1.430} & 0.900 & 2.857 & 1.157 & 2.770 & 1.125 & 1.470 & 0.788 \\
\bottomrule
\end{tabular}
\end{table}

\newpage

\section{Short-term Forecasting Comparison with Transformer Models}
The detailed comparison of ForecastGAN with transformer models is presented in Table \ref{tbl:short_results}. The look-back window $S$ is kept at 12 for all datasets except for the illness dataset with $S=2$.
 
\begin{sidewaystable}
\centering
\caption{Performance comparison of different models for short-term forecasting}
\label{tbl:short_results}
\setlength{\tabcolsep}{2pt} 
\begin{tabular*}{\textheight}{@{\extracolsep\fill}l*{20}{c}@{}}
\toprule
\multicolumn{2}{c}{\multirow{1}{*}{\textbf{Methods}}} & 
\multicolumn{2}{c}{\multirow{1}{*}{\textbf{ForecastGAN*}}} & 
\multicolumn{2}{c}{\multirow{1}{*}{\textbf{FEDformer}}} & 
\multicolumn{2}{c}{\multirow{1}{*}{\textbf{Autoformer}}} & 
\multicolumn{2}{c}{\multirow{1}{*}{\textbf{Informer}}} & 
\multicolumn{2}{c}{\multirow{1}{*}{\textbf{Crossformer}}} &
\multicolumn{2}{c}{\multirow{1}{*}{\textbf{PatchTST}}} & 
\multicolumn{2}{c}{\multirow{1}{*}{\textbf{Client}}} 
\\ 
\midrule
\textbf{Data} & \textbf{H} & \textbf{MSE} & \textbf{MAE} & \textbf{MSE} & \textbf{MAE} & \textbf{MSE} & \textbf{MAE} & \textbf{MSE} & \textbf{MAE} & \textbf{MSE} & \textbf{MAE} & \textbf{MSE} & \textbf{MAE} & \textbf{MSE} & \textbf{MAE} \\
\midrule
Exchange  & 12 & \textbf{0.196} & 0.071 & \underline{0.46} & 0.415 & 0.984 & 0.334 & 0.984 & 0.889 & 1.598 & 0.425 & 1.102 & 0.597 & 1.735 & 0.734 \\
  & 24 & \textbf{0.288} & 0.138 & \underline{0.506} & 0.517 & 1.341 & 0.437 & 1.341 & 1.032 & 1.432 & 0.395 & 0.791 & 0.643 & 1.569 & 0.780 \\
  & 32 & \textbf{0.397} & 0.281 & \underline{0.661} & 0.637 & 1.809 & 0.646 & 1.809 & 1.173 & 1.389 & 0.386 & 1.124 & 0.798 & 1.526 & 0.935 \\
  & 48 & \textbf{0.716} & 0.625 & \underline{1.078} & 0.978 & 2.615 & 1.584 & 2.615 & 1.447 & 1.31 & 0.369 & 1.227 & 1.215 & 1.447 & 1.352 \\
\hline

Electricity & 12 & \textbf{0.21} & 0.121 & 0.464 & 0.349 & 0.473 & 0.357 & 0.43 & 0.524 & 0.459 & 0.412 & \underline{0.378} & 0.620 & 0.615 & 0.776 \\
  & 24 & \textbf{0.141} & 0.138 & 0.49 & 0.471 & 0.452 & 0.378 & 0.452 & 0.542 & 0.45 & 0.382 & \underline{0.396} & 0.646 & 0.606 & 0.802 \\
  & 32 & \textbf{0.243} & 0.151 & 0.494 & 0.485 & 0.456 & 0.387 & 0.456 & 0.550 & 0.423 & 0.373 & \underline{0.415} & 0.650 & 0.579 & 0.806 \\
  & 48 & \textbf{0.299} & 0.191 & 0.517 & 0.511 & 0.529 & 0.410 & 0.529 & 0.595 & \underline{0.439} & 0.356 & 0.446 & 0.673 & 0.595 & 0.829 \\
\hline

ETTh1  & 12 & \textbf{0.279} & 0.332 & 0.536 & 0.493 & 0.576 & 0.566 & 0.982 & 0.830 & 0.562 & 0.509 & \underline{0.517} & 0.653 & 0.679 & 0.770 \\
  & 24 & \textbf{0.27} & 0.310 & 0.565 & 0.537 & 0.599 & 0.617 & 1.125 & 0.909 & \underline{0.532} & 0.500 & 0.546 & 0.682 & 0.649 & 0.799 \\
  & 32 & \textbf{0.243} & 0.302 & 0.582 & 0.576 & 0.613 & 0.638 & 1.224 & 0.926 & \underline{0.523} & 0.473 & 0.557 & 0.699 & 0.64 & 0.816 \\
  & 48 & \textbf{0.251} & 0.340 & 0.624 & 0.623 & 0.629 & 0.631 & 1.298 & 0.982 & \underline{0.506} & 0.481 & 0.585 & 0.741 & 0.623 & 0.858 \\
\hline

ETTm1  & 12 & \textbf{0.157} & 0.167 & 0.528 & 0.488 & 0.584 & 0.614 & 0.781 & 0.680 & \underline{0.404} & 0.387 & 0.455 & 0.637 & 0.513 & 0.746 \\
  & 24 & \textbf{0.168} & 0.175 & 0.55 & 0.535 & 0.605 & 0.662 & 0.904 & 0.778 & \underline{0.413} & 0.398 & 0.479 & 0.659 & 0.522 & 0.768 \\
  & 32 & \textbf{0.26} & 0.354 & 0.568 & 0.554 & 0.646 & 0.730 & 1.321 & 0.980 & 0.584 & 0.490 & \underline{0.501} & 0.677 & 0.693 & 0.786 \\
  & 48 & \textbf{0.293} & 0.387 & 0.599 & 0.652 & 0.67 & 0.780 & 1.275 & 0.932 & 0.617 & 0.523 & \underline{0.529} & 0.708 & 0.726 & 0.817 \\
\hline

Traffic & 12 & \textbf{0.257} & 0.356 & 0.468 & 0.727 & 0.49 & 0.753 & 0.859 & 0.493 & 0.987 & 0.674 & \underline{0.351} & 0.570 & 1.127 & 0.672 \\
  & 24 & \textbf{0.285} & 0.395 & 0.718 & 0.513 & 0.798 & 0.522 & 0.519 & 0.969 & 0.753 & 0.446 & \underline{0.358} & 0.820 & 0.893 & 0.922 \\
  & 32 & \textbf{0.293} & 0.401 & 0.485 & 0.761 & 0.439 & 0.762 & 0.917 & 0.522 & 0.699 & 0.432 & \underline{0.366} & 0.587 & 0.839 & 0.689 \\
  & 48 & \textbf{0.301} & 0.428 & 0.484 & 0.766 & 0.51 & 0.800 & 1.004 & 0.574 & 0.601 & 0.398 & \underline{0.388} & 0.586 & 0.741 & 0.688 \\
\hline

Weather  & 12 & \textbf{0.198} & 0.145 & 0.399 & 0.320 & 0.439 & 0.369 & 0.403 & 0.487 & 0.722 & 0.669 & \underline{0.301} & 0.502 & 0.825 & 0.605 \\
 & 24 & \textbf{0.216} & 0.178 & 0.439 & 0.379 & 0.47 & 0.410 & 0.701 & 0.647 & 0.692 & 0.660 & \underline{0.344} & 0.542 & 0.795 & 0.645 \\
 & 32 & \textbf{0.268} & 0.218 & 0.483 & 0.442 & 0.498 & 0.462 & 0.681 & 0.626 & 0.683 & 0.633 & \underline{0.385} & 0.586 & 0.786 & 0.689 \\
 & 48 & \textbf{0.311} & 0.281 & 0.531 & 0.506 & 0.531 & 0.522 & 1.162 & 0.844 & 0.666 & 0.641 & \underline{0.437} & 0.634 & 0.769 & 0.737 \\
\hline

illness  & 2 & \textbf{0.854} & 1.661 & 1.497 & 3.465 & 1.524 & 3.720 & 6.001 & 1.914 & 2.161 & 1.086 & \underline{0.991} & 1.734 & 2.398 & 1.971 \\
 & 6 & \textbf{0.857} & 1.692 & 1.317 & 2.916 & 1.385 & 3.340 & 4.992 & 1.704 & 2.307 & 1.097 & \underline{1.107} & 1.554 & 2.544 & 1.791 \\
 & 8 & \textbf{0.878} & 1.721 & 1.315 & 2.859 & 1.322 & 2.906 & 5 & 1.706 & 2.389 & 1.106 & \underline{1.052} & 1.552 & 2.626 & 1.789 \\
 & 10 & \textbf{0.9} & 1.803 & 1.394 & 3.094 & 1.362 & 3.007 & 5.501 & 1.801 & 2.431 & 1.180 & \underline{1.025} & 1.631 & 2.668 & 1.868 \\

\bottomrule
\end{tabular*}
\end{sidewaystable}

\end{appendices}
\newpage


 \bibliographystyle{elsarticle-num} 

\bibliography{sn-bibliography}

\begin{thebibliography}{10}
\expandafter\ifx\csname url\endcsname\relax
  \def\url#1{\texttt{#1}}\fi
\expandafter\ifx\csname urlprefix\endcsname\relax\def\urlprefix{URL }\fi
\expandafter\ifx\csname href\endcsname\relax
  \def\href#1#2{#2} \def\path#1{#1}\fi

\bibitem{Miller2021}
D.~Miller, J.-M. Kim, \href{http://dx.doi.org/10.3390/jrfm14100486}{Univariate and multivariate machine learning forecasting models on the price returns of cryptocurrencies}, Journal of Risk and Financial Management 14~(10) (2021) 486.
\newblock \href {https://doi.org/10.3390/jrfm14100486} {\path{doi:10.3390/jrfm14100486}}.
\newline\urlprefix\url{http://dx.doi.org/10.3390/jrfm14100486}

\bibitem{Hewage2020}
P.~Hewage, M.~Trovati, E.~Pereira, A.~Behera, \href{http://dx.doi.org/10.1007/s10044-020-00898-1}{Deep learning-based effective fine-grained weather forecasting model}, Pattern Analysis and Applications 24~(1) (2020) 343–366.
\newblock \href {https://doi.org/10.1007/s10044-020-00898-1} {\path{doi:10.1007/s10044-020-00898-1}}.
\newline\urlprefix\url{http://dx.doi.org/10.1007/s10044-020-00898-1}

\bibitem{Punyapornwithaya2023}
V.~Punyapornwithaya, O.~Arjkumpa, N.~Buamithup, N.~Kuatako, K.~Klaharn, C.~Sansamur, K.~Jampachaisri, \href{http://dx.doi.org/10.1016/j.prevetmed.2023.105964}{Forecasting of daily new lumpy skin disease cases in thailand at different stages of the epidemic using fuzzy logic time series, nnar, and arima methods}, Preventive Veterinary Medicine 217 (2023) 105964.
\newblock \href {https://doi.org/10.1016/j.prevetmed.2023.105964} {\path{doi:10.1016/j.prevetmed.2023.105964}}.
\newline\urlprefix\url{http://dx.doi.org/10.1016/j.prevetmed.2023.105964}

\bibitem{Abbasimehr2023}
H.~Abbasimehr, R.~Paki, A.~Bahrini, \href{http://dx.doi.org/10.1016/j.suscom.2023.100863}{A novel xgboost-based featurization approach to forecast renewable energy consumption with deep learning models}, Sustainable Computing: Informatics and Systems 38 (2023) 100863.
\newblock \href {https://doi.org/10.1016/j.suscom.2023.100863} {\path{doi:10.1016/j.suscom.2023.100863}}.
\newline\urlprefix\url{http://dx.doi.org/10.1016/j.suscom.2023.100863}

\bibitem{randfor}
B.~Goehry, H.~Yan, Y.~Goude, P.~Massart, J.-M. Poggi, Random forests for time series, REVSTAT-Statistical Journal (2023) Vol. 21 No. 2 (2023): REVSTAT--Statistical Journal\href {https://doi.org/10.57805/REVSTAT.V21I2.400} {\path{doi:10.57805/REVSTAT.V21I2.400}}.

\bibitem{Kurani2021}
A.~Kurani, P.~Doshi, A.~Vakharia, M.~Shah, \href{http://dx.doi.org/10.1007/s40745-021-00344-x}{A comprehensive comparative study of artificial neural network (ann) and support vector machines (svm) on stock forecasting}, Annals of Data Science 10~(1) (2021) 183–208.
\newblock \href {https://doi.org/10.1007/s40745-021-00344-x} {\path{doi:10.1007/s40745-021-00344-x}}.
\newline\urlprefix\url{http://dx.doi.org/10.1007/s40745-021-00344-x}

\bibitem{Dudukcu2023}
H.~V. Dudukcu, M.~Taskiran, Z.~G. Cam~Taskiran, T.~Yildirim, \href{http://dx.doi.org/10.1016/j.asoc.2022.109945}{Temporal convolutional networks with rnn approach for chaotic time series prediction}, Applied Soft Computing 133 (2023) 109945.
\newblock \href {https://doi.org/10.1016/j.asoc.2022.109945} {\path{doi:10.1016/j.asoc.2022.109945}}.
\newline\urlprefix\url{http://dx.doi.org/10.1016/j.asoc.2022.109945}

\bibitem{cnn}
Z.~Zeng, R.~Kaur, S.~Siddagangappa, S.~Rahimi, T.~Balch, M.~Veloso, \href{https://arxiv.org/abs/2304.04912}{Financial time series forecasting using cnn and transformer} (2023).
\newblock \href {https://doi.org/10.48550/ARXIV.2304.04912} {\path{doi:10.48550/ARXIV.2304.04912}}.
\newline\urlprefix\url{https://arxiv.org/abs/2304.04912}

\bibitem{goodfellow2014generative}
I.~Goodfellow, J.~Pouget-Abadie, M.~Mirza, B.~Xu, D.~Warde-Farley, S.~Ozair, A.~Courville, Y.~Bengio, Generative adversarial nets, Advances in neural information processing systems 27 (2014).

\bibitem{vaswani2017attention}
A.~Vaswani, N.~Shazeer, N.~Parmar, J.~Uszkoreit, L.~Jones, A.~N. Gomez, {\L}.~Kaiser, I.~Polosukhin, Attention is all you need, Advances in neural information processing systems 30 (2017).

\bibitem{zeng2023transformers}
A.~Zeng, M.~Chen, L.~Zhang, Q.~Xu, Are transformers effective for time series forecasting?, in: Proceedings of the AAAI conference on artificial intelligence, Vol.~37, 2023, pp. 11121--11128.

\bibitem{WishalFatima2023}
S.~S.~W. Fatima, A.~Rahimi, K.~Hayat, Comparison of prophet with machine learning regression models for long and short-term manufacturing forecasting, in: 2023 28th International Conference on Automation and Computing (ICAC), IEEE, 2023, p.~1.
\newblock \href {https://doi.org/10.1109/icac57885.2023.10275201} {\path{doi:10.1109/icac57885.2023.10275201}}.

\bibitem{hyndman2018}
R.~J. Hyndman, G.~Athanasopoulos, Forecasting: principles and practice, 3rd Edition, OTexts, Australia, 2021.

\bibitem{gardner2006exponential}
E.~S. Gardner, \href{http://dx.doi.org/10.1016/j.ijforecast.2006.03.005}{Exponential smoothing: The state of the art—part ii}, International Journal of Forecasting 22~(4) (2006) 637–666.
\newblock \href {https://doi.org/10.1016/j.ijforecast.2006.03.005} {\path{doi:10.1016/j.ijforecast.2006.03.005}}.
\newline\urlprefix\url{http://dx.doi.org/10.1016/j.ijforecast.2006.03.005}

\bibitem{mahaseth2022short}
R.~Mahaseth, N.~Kumar, A.~Aggarwal, A.~Tayal, A.~Kumar, R.~Gupta, \href{http://dx.doi.org/10.1080/02522667.2022.2042093}{Short term wind power forecasting using k-nearest neighbour (knn)}, Journal of Information and Optimization Sciences 43~(1) (2022) 251–259.
\newblock \href {https://doi.org/10.1080/02522667.2022.2042093} {\path{doi:10.1080/02522667.2022.2042093}}.
\newline\urlprefix\url{http://dx.doi.org/10.1080/02522667.2022.2042093}

\bibitem{lecun2015deep}
Y.~LeCun, Y.~Bengio, G.~Hinton, \href{http://dx.doi.org/10.1038/nature14539}{Deep learning}, Nature 521~(7553) (2015) 436–444.
\newblock \href {https://doi.org/10.1038/nature14539} {\path{doi:10.1038/nature14539}}.
\newline\urlprefix\url{http://dx.doi.org/10.1038/nature14539}

\bibitem{kamarianakis2012real}
Y.~Kamarianakis, W.~Shen, L.~Wynter, \href{http://dx.doi.org/10.1002/asmb.1937}{Real‐time road traffic forecasting using regime‐switching space‐time models and adaptive lasso}, Applied Stochastic Models in Business and Industry 28~(4) (2012) 297–315.
\newblock \href {https://doi.org/10.1002/asmb.1937} {\path{doi:10.1002/asmb.1937}}.
\newline\urlprefix\url{http://dx.doi.org/10.1002/asmb.1937}

\bibitem{Fatima2024}
S.~S.~W. Fatima, A.~Rahimi, \href{http://dx.doi.org/10.3390/machines12060380}{A review of time-series forecasting algorithms for industrial manufacturing systems}, Machines 12~(6) (2024) 380.
\newblock \href {https://doi.org/10.3390/machines12060380} {\path{doi:10.3390/machines12060380}}.
\newline\urlprefix\url{http://dx.doi.org/10.3390/machines12060380}

\bibitem{wiese2020quant}
M.~Wiese, R.~Knobloch, R.~Korn, P.~Kretschmer, \href{http://dx.doi.org/10.1080/14697688.2020.1730426}{Quant gans: deep generation of financial time series}, Quantitative Finance 20~(9) (2020) 1419–1440.
\newblock \href {https://doi.org/10.1080/14697688.2020.1730426} {\path{doi:10.1080/14697688.2020.1730426}}.
\newline\urlprefix\url{http://dx.doi.org/10.1080/14697688.2020.1730426}

\bibitem{koochali2019probabilistic}
A.~Koochali, P.~Schichtel, A.~Dengel, S.~Ahmed, Probabilistic forecasting of sensory data with generative adversarial networks – forgan, IEEE Access 7 (2019) 63868--63880.
\newblock \href {https://doi.org/10.1109/ACCESS.2019.2915544} {\path{doi:10.1109/ACCESS.2019.2915544}}.

\bibitem{zhang2019stock}
K.~Zhang, G.~Zhong, J.~Dong, S.~Wang, Y.~Wang, \href{http://dx.doi.org/10.1016/j.procs.2019.01.256}{Stock market prediction based on generative adversarial network}, Procedia Computer Science 147 (2019) 400–406.
\newblock \href {https://doi.org/10.1016/j.procs.2019.01.256} {\path{doi:10.1016/j.procs.2019.01.256}}.
\newline\urlprefix\url{http://dx.doi.org/10.1016/j.procs.2019.01.256}

\bibitem{zhou2018stock}
X.~Zhou, Z.~Pan, G.~Hu, S.~Tang, C.~Zhao, \href{http://dx.doi.org/10.1155/2018/4907423}{Stock market prediction on high-frequency data using generative adversarial nets}, Mathematical Problems in Engineering 2018 (2018) 1–11.
\newblock \href {https://doi.org/10.1155/2018/4907423} {\path{doi:10.1155/2018/4907423}}.
\newline\urlprefix\url{http://dx.doi.org/10.1155/2018/4907423}

\bibitem{lin2018pattern}
Y.~Lin, X.~Dai, L.~Li, F.-Y. Wang, Pattern sensitive prediction of traffic flow based on generative adversarial framework, IEEE Transactions on Intelligent Transportation Systems 20~(6) (2019) 2395--2400.
\newblock \href {https://doi.org/10.1109/TITS.2018.2857224} {\path{doi:10.1109/TITS.2018.2857224}}.

\bibitem{Harvey1990}
A.~C. Harvey, S.~Peters, \href{http://dx.doi.org/10.1002/for.3980090203}{Estimation procedures for structural time series models}, Journal of Forecasting 9~(2) (1990) 89–108.
\newblock \href {https://doi.org/10.1002/for.3980090203} {\path{doi:10.1002/for.3980090203}}.
\newline\urlprefix\url{http://dx.doi.org/10.1002/for.3980090203}

\bibitem{wu2021autoformer}
H.~Wu, J.~Xu, J.~Wang, M.~Long, Autoformer: Decomposition transformers with auto-correlation for long-term series forecasting, Advances in neural information processing systems 34 (2021) 22419--22430.

\bibitem{taylor2018forecasting}
S.~J. Taylor, B.~Letham, Forecasting at scale, The American Statistician 72~(1) (2018) 37--45.

\bibitem{neural_forecasting}
K.~Benidis, S.~S. Rangapuram, V.~Flunkert, B.~Wang, D.~C. Maddix, A.~C. T{\"{u}}rkmen, J.~Gasthaus, M.~Bohlke{-}Schneider, D.~Salinas, L.~Stella, L.~Callot, T.~Januschowski, \href{https://arxiv.org/abs/2004.10240}{Neural forecasting: Introduction and literature overview}, CoRR abs/2004.10240 (2020).
\newblock \href {http://arxiv.org/abs/2004.10240} {\path{arXiv:2004.10240}}.
\newline\urlprefix\url{https://arxiv.org/abs/2004.10240}

\bibitem{nbeats}
B.~N. Oreshkin, D.~Carpov, N.~Chapados, Y.~Bengio, \href{https://arxiv.org/abs/1905.10437}{N-beats: Neural basis expansion analysis for interpretable time series forecasting} (2019).
\newblock \href {https://doi.org/10.48550/ARXIV.1905.10437} {\path{doi:10.48550/ARXIV.1905.10437}}.
\newline\urlprefix\url{https://arxiv.org/abs/1905.10437}

\bibitem{NEURIPS2019_3a0844ce}
R.~Sen, H.-F. Yu, I.~Dhillon, \href{https://arxiv.org/abs/1905.03806}{Think globally, act locally: A deep neural network approach to high-dimensional time series forecasting} (2019).
\newblock \href {https://doi.org/10.48550/ARXIV.1905.03806} {\path{doi:10.48550/ARXIV.1905.03806}}.
\newline\urlprefix\url{https://arxiv.org/abs/1905.03806}

\bibitem{trans}
M.~Liu, A.~Zeng, M.~Chen, Z.~Xu, Q.~Lai, L.~Ma, Q.~Xu, \href{https://arxiv.org/abs/2106.09305}{Scinet: Time series modeling and forecasting with sample convolution and interaction} (2021).
\newblock \href {https://doi.org/10.48550/ARXIV.2106.09305} {\path{doi:10.48550/ARXIV.2106.09305}}.
\newline\urlprefix\url{https://arxiv.org/abs/2106.09305}

\bibitem{nguyen2022improving}
T.~M. Nguyen, T.~M. Nguyen, D.~D. Le, D.~K. Nguyen, V.-A. Tran, R.~Baraniuk, N.~Ho, S.~Osher, Improving transformers with probabilistic attention keys, in: International Conference on Machine Learning, PMLR, 2022, pp. 16595--16621.

\bibitem{NEURIPS2022_266983d0}
M.~LIU, A.~Zeng, M.~Chen, Z.~Xu, Q.~LAI, L.~Ma, Q.~Xu, Scinet: Time series modeling and forecasting with sample convolution and interaction, in: S.~Koyejo, S.~Mohamed, A.~Agarwal, D.~Belgrave, K.~Cho, A.~Oh (Eds.), Advances in Neural Information Processing Systems, Vol.~35, Curran Associates, Inc., 2022, pp. 5816--5828.

\bibitem{Zhao2022}
W.~Zhao, S.~Alwidian, Q.~H. Mahmoud, \href{http://dx.doi.org/10.3390/a15080283}{Adversarial training methods for deep learning: A systematic review}, Algorithms 15~(8) (2022) 283.
\newblock \href {https://doi.org/10.3390/a15080283} {\path{doi:10.3390/a15080283}}.
\newline\urlprefix\url{http://dx.doi.org/10.3390/a15080283}

\bibitem{Koochali2021}
A.~Koochali, A.~Dengel, S.~Ahmed, If you like it, gan it—probabilistic multivariate times series forecast with gan, in: The 7th International Conference on Time Series and Forecasting, ITISE 2021, MDPI, 2021, p.~40.
\newblock \href {https://doi.org/10.3390/engproc2021005040} {\path{doi:10.3390/engproc2021005040}}.

\bibitem{terzi2021adversarial}
M.~Terzi, A.~Achille, M.~Maggipinto, G.~A. Susto, Adversarial training reduces information and improves transferability, in: Proceedings of the AAAI Conference on Artificial Intelligence, Vol.~35, 2021, pp. 2674--2682.

\bibitem{haoyietal-informer-2021}
H.~Zhou, S.~Zhang, J.~Peng, S.~Zhang, J.~Li, H.~Xiong, W.~Zhang, Informer: Beyond efficient transformer for long sequence time-series forecasting, Proc. Conf. AAAI Artif. Intell. 35~(12) (2021) 11106--11115.

\bibitem{data_prod}
A.~Trindade, {Productivity Prediction of Garment Employees}, UCI Machine Learning Repository, {DOI}: https://doi.org/10.24432/C51S6D (2020).

\bibitem{misc_electricityloaddiagrams20112014_321}
A.~Trindade, {ElectricityLoadDiagrams20112014}, UCI Machine Learning Repository, {DOI}: https://doi.org/10.24432/C58C86 (2015).

\bibitem{illness_data}
CDC, \href{gis.cdc.gov/grasp/fluview/fluportaldashboard.html}{National, regional, and state level outpatient illness and viral surveillance} (2024).
\newline\urlprefix\url{gis.cdc.gov/grasp/fluview/fluportaldashboard.html}

\bibitem{exchange_rate_data}
G.~Lai, \href{github.com/laiguokun/multivariate-time-series-data}{Multivariate-time-series-data}, Github (2017).
\newline\urlprefix\url{github.com/laiguokun/multivariate-time-series-data}

\bibitem{weather_data}
O.~Kolle, \href{https://www.bgc-jena.mpg.de/wetter/}{Weather station saaleaue} (2024).
\newline\urlprefix\url{https://www.bgc-jena.mpg.de/wetter/}

\bibitem{stock_data}
UCI, \href{https://archive.ics.uci.edu/dataset/554}{Cnnpred: Cnn-based stock market prediction using a diverse set of variables}, UCI Machine Learning Repository (2019).
\newblock \href {https://doi.org/10.24432/C55P70} {\path{doi:10.24432/C55P70}}.
\newline\urlprefix\url{https://archive.ics.uci.edu/dataset/554}

\bibitem{robformer}
Y.~Yu, R.~Ma, Z.~Ma, \href{https://www.sciencedirect.com/science/article/pii/S0031320324003030}{Robformer: A robust decomposition transformer for long-term time series forecasting}, Pattern Recognition 153 (2024) 110552.
\newblock \href {https://doi.org/https://doi.org/10.1016/j.patcog.2024.110552} {\path{doi:https://doi.org/10.1016/j.patcog.2024.110552}}.
\newline\urlprefix\url{https://www.sciencedirect.com/science/article/pii/S0031320324003030}

\bibitem{timexer}
Y.~Wang, H.~Wu, J.~Dong, G.~Qin, H.~Zhang, Y.~Liu, Y.~Qiu, J.~Wang, M.~Long, \href{https://arxiv.org/abs/2402.19072}{Timexer: Empowering transformers for time series forecasting with exogenous variables} (2024).
\newblock \href {http://arxiv.org/abs/2402.19072} {\path{arXiv:2402.19072}}.
\newline\urlprefix\url{https://arxiv.org/abs/2402.19072}

\bibitem{zhang2022crossformer}
Y.~Zhang, J.~Yan, Crossformer: Transformer utilizing cross-dimension dependency for multivariate time series forecasting, in: The eleventh international conference on learning representations, 2022, p.~1.

\bibitem{pathformer}
P.~Chen, Y.~Zhang, Y.~Cheng, Y.~Shu, Y.~Wang, Q.~Wen, B.~Yang, C.~Guo, \href{https://arxiv.org/abs/2402.05956}{Pathformer: Multi-scale transformers with adaptive pathways for time series forecasting} (2024).
\newblock \href {http://arxiv.org/abs/2402.05956} {\path{arXiv:2402.05956}}.
\newline\urlprefix\url{https://arxiv.org/abs/2402.05956}

\bibitem{client}
J.~Gao, W.~Hu, Y.~Chen, \href{https://arxiv.org/abs/2305.18838}{Client: Cross-variable linear integrated enhanced transformer for multivariate long-term time series forecasting} (2023).
\newblock \href {https://doi.org/10.48550/ARXIV.2305.18838} {\path{doi:10.48550/ARXIV.2305.18838}}.
\newline\urlprefix\url{https://arxiv.org/abs/2305.18838}

\bibitem{8765895}
J.~Wang, Z.~Yang, J.~Zhang, Q.~Zhang, W.-T.~K. Chien, Adabalgan: An improved generative adversarial network with imbalanced learning for wafer defective pattern recognition, IEEE Transactions on Semiconductor Manufacturing 32~(3) (2019) 310--319.
\newblock \href {https://doi.org/10.1109/TSM.2019.2925361} {\path{doi:10.1109/TSM.2019.2925361}}.

\bibitem{9592834}
Y.~Li, Z.~Shi, C.~Liu, W.~Tian, Z.~Kong, C.~B. Williams, Augmented time regularized generative adversarial network (atr-gan) for data augmentation in online process anomaly detection, IEEE Transactions on Automation Science and Engineering 19~(4) (2022) 3338--3355.
\newblock \href {https://doi.org/10.1109/TASE.2021.3118635} {\path{doi:10.1109/TASE.2021.3118635}}.

\bibitem{9456937}
J.~Kim, Y.~Nam, M.-C. Kang, K.~Kim, J.~Hong, S.~Lee, D.-N. Kim, Adversarial defect detection in semiconductor manufacturing process, IEEE Transactions on Semiconductor Manufacturing 34~(3) (2021) 365--371.
\newblock \href {https://doi.org/10.1109/TSM.2021.3089869} {\path{doi:10.1109/TSM.2021.3089869}}.

\bibitem{gao2022deep}
H.~Gao, Y.~Zhang, W.~Lv, J.~Yin, T.~Qasim, D.~Wang, \href{http://dx.doi.org/10.3390/app12136569}{A deep convolutional generative adversarial networks-based method for defect detection in small sample industrial parts images}, Applied Sciences 12~(13) (2022) 6569.
\newblock \href {https://doi.org/10.3390/app12136569} {\path{doi:10.3390/app12136569}}.
\newline\urlprefix\url{http://dx.doi.org/10.3390/app12136569}

\bibitem{luo2021case}
J.~Luo, J.~Huang, H.~Li, \href{http://dx.doi.org/10.1007/s10845-020-01579-w}{A case study of conditional deep convolutional generative adversarial networks in machine fault diagnosis}, Journal of Intelligent Manufacturing 32~(2) (2020) 407–425.
\newblock \href {https://doi.org/10.1007/s10845-020-01579-w} {\path{doi:10.1007/s10845-020-01579-w}}.
\newline\urlprefix\url{http://dx.doi.org/10.1007/s10845-020-01579-w}

\bibitem{mumbelli2023application}
J.~D. Mumbelli, G.~A. Guarneri, Y.~K. Lopes, D.~Casanova, M.~Teixeira, \href{http://dx.doi.org/10.1016/j.asoc.2023.110105}{An application of generative adversarial networks to improve automatic inspection in automotive manufacturing}, Applied Soft Computing 136 (2023) 110105.
\newblock \href {https://doi.org/10.1016/j.asoc.2023.110105} {\path{doi:10.1016/j.asoc.2023.110105}}.
\newline\urlprefix\url{http://dx.doi.org/10.1016/j.asoc.2023.110105}

\bibitem{Sun2021}
S.~Sun, X.~Hu, Y.~Liu, \href{http://dx.doi.org/10.1007/s10845-021-01806-y}{An imbalanced data learning method for tool breakage detection based on generative adversarial networks}, Journal of Intelligent Manufacturing 33~(8) (2021) 2441–2455.
\newblock \href {https://doi.org/10.1007/s10845-021-01806-y} {\path{doi:10.1007/s10845-021-01806-y}}.
\newline\urlprefix\url{http://dx.doi.org/10.1007/s10845-021-01806-y}

\bibitem{Cooper2020}
C.~Cooper, J.~Zhang, R.~X. Gao, P.~Wang, I.~Ragai, \href{http://dx.doi.org/10.1016/j.promfg.2020.05.059}{Anomaly detection in milling tools using acoustic signals and generative adversarial networks}, Procedia Manufacturing 48 (2020) 372–378.
\newblock \href {https://doi.org/10.1016/j.promfg.2020.05.059} {\path{doi:10.1016/j.promfg.2020.05.059}}.
\newline\urlprefix\url{http://dx.doi.org/10.1016/j.promfg.2020.05.059}

\bibitem{9151342}
J.~Dai, J.~Wang, W.~Huang, J.~Shi, Z.~Zhu, Machinery health monitoring based on unsupervised feature learning via generative adversarial networks, IEEE/ASME Transactions on Mechatronics 25~(5) (2020) 2252--2263.
\newblock \href {https://doi.org/10.1109/TMECH.2020.3012179} {\path{doi:10.1109/TMECH.2020.3012179}}.

\bibitem{Hoh2022}
M.~Hoh, A.~Sch\"{o}ttl, H.~Schaub, F.~Wenninger, \href{http://dx.doi.org/10.1016/j.procs.2022.01.261}{A generative model for anomaly detection in time series data}, Procedia Computer Science 200 (2022) 629–637.
\newblock \href {https://doi.org/10.1016/j.procs.2022.01.261} {\path{doi:10.1016/j.procs.2022.01.261}}.
\newline\urlprefix\url{http://dx.doi.org/10.1016/j.procs.2022.01.261}

\bibitem{9308512}
M.~A. Bashar, R.~Nayak, Tanogan: Time series anomaly detection with generative adversarial networks, in: 2020 IEEE Symposium Series on Computational Intelligence (SSCI), 2020, pp. 1778--1785.
\newblock \href {https://doi.org/10.1109/SSCI47803.2020.9308512} {\path{doi:10.1109/SSCI47803.2020.9308512}}.

\bibitem{Song2022}
J.~Song, Y.~C. Lee, J.~Lee, \href{http://dx.doi.org/10.1007/s10845-022-01981-6}{Deep generative model with time series-image encoding for manufacturing fault detection in die casting process}, Journal of Intelligent Manufacturing 34~(7) (2022) 3001–3014.
\newblock \href {https://doi.org/10.1007/s10845-022-01981-6} {\path{doi:10.1007/s10845-022-01981-6}}.
\newline\urlprefix\url{http://dx.doi.org/10.1007/s10845-022-01981-6}

\bibitem{Balzategui2021}
J.~Balzategui, L.~Eciolaza, D.~Maestro-Watson, \href{http://dx.doi.org/10.3390/s21134361}{Anomaly detection and automatic labeling for solar cell quality inspection based on generative adversarial network}, Sensors 21~(13) (2021) 4361.
\newblock \href {https://doi.org/10.3390/s21134361} {\path{doi:10.3390/s21134361}}.
\newline\urlprefix\url{http://dx.doi.org/10.3390/s21134361}

\bibitem{Bazarbaev2021}
M.~Bazarbaev, T.~Chuluunsaikhan, H.~Oh, G.-A. Ryu, A.~Nasridinov, K.-H. Yoo, \href{http://dx.doi.org/10.3390/s22010029}{Generation of time-series working patterns for manufacturing high-quality products through auxiliary classifier generative adversarial network}, Sensors 22~(1) (2021) 29.
\newblock \href {https://doi.org/10.3390/s22010029} {\path{doi:10.3390/s22010029}}.
\newline\urlprefix\url{http://dx.doi.org/10.3390/s22010029}

\bibitem{8850773}
Y.~Wang, K.~Li, S.~Gan, C.~Cameron, M.~Zheng, Data augmentation for intelligent manufacturing with generative adversarial framework, in: 2019 1st International Conference on Industrial Artificial Intelligence (IAI), 2019, pp. 1--6.
\newblock \href {https://doi.org/10.1109/ICIAI.2019.8850773} {\path{doi:10.1109/ICIAI.2019.8850773}}.

\bibitem{Du2022}
W.~Du, Z.~Xia, L.~Han, B.~Gao, \href{http://dx.doi.org/10.1007/s10489-022-04381-8}{3d solid model generation method based on a generative adversarial network}, Applied Intelligence 53~(13) (2022) 17035–17060.
\newblock \href {https://doi.org/10.1007/s10489-022-04381-8} {\path{doi:10.1007/s10489-022-04381-8}}.
\newline\urlprefix\url{http://dx.doi.org/10.1007/s10489-022-04381-8}

\bibitem{Ye2019}
W.~Ye, M.~B. Alawieh, Y.~Lin, D.~Z. Pan, \href{http://dx.doi.org/10.1145/3316781.3317852}{Lithogan: End-to-end lithography modeling with generative adversarial networks}, in: Proceedings of the 56th Annual Design Automation Conference 2019, DAC ’19, ACM, 2019, p.~1.
\newblock \href {https://doi.org/10.1145/3316781.3317852} {\path{doi:10.1145/3316781.3317852}}.
\newline\urlprefix\url{http://dx.doi.org/10.1145/3316781.3317852}

\bibitem{gobert2019conditional}
C.~Gobert, E.~Arrieta, A.~Belmontes, R.~B. Wicker, F.~Medina, B.~McWilliams, \href{https://repositories.lib.utexas.edu/handle/2152/90329}{Conditional generative adversarial networks for in-situ layerwise additive manufacturing data} (2019).
\newblock \href {https://doi.org/10.26153/TSW/17250} {\path{doi:10.26153/TSW/17250}}.
\newline\urlprefix\url{https://repositories.lib.utexas.edu/handle/2152/90329}

\bibitem{Wang2021}
Q.~Wang, R.~Yang, C.~Wu, Y.~Liu, \href{http://dx.doi.org/10.1016/j.jmapro.2021.03.053}{An effective defect detection method based on improved generative adversarial networks (igan) for machined surfaces}, Journal of Manufacturing Processes 65 (2021) 373–381.
\newblock \href {https://doi.org/10.1016/j.jmapro.2021.03.053} {\path{doi:10.1016/j.jmapro.2021.03.053}}.
\newline\urlprefix\url{http://dx.doi.org/10.1016/j.jmapro.2021.03.053}

\bibitem{yan2023automated}
X.~Yan, S.~Melkote, Automated manufacturability analysis and machining process selection using deep generative model and siamese neural networks, Journal of Manufacturing Systems 67 (2023) 57--67.
\newblock \href {https://doi.org/https://doi.org/10.1016/j.jmsy.2023.01.006} {\path{doi:https://doi.org/10.1016/j.jmsy.2023.01.006}}.

\bibitem{liu2021melt}
W.~Liu, Z.~Wang, L.~Tian, S.~Lauria, X.~Liu, Melt pool segmentation for additive manufacturing: A generative adversarial network approach, Computers and Electrical Engineering 92 (2021) 107183.
\newblock \href {https://doi.org/https://doi.org/10.1016/j.compeleceng.2021.107183} {\path{doi:https://doi.org/10.1016/j.compeleceng.2021.107183}}.

\bibitem{greminger2020}
M.~Greminger, \href{https://doi.org/10.1115/DETC2020-22399}{{Generative Adversarial Networks With Synthetic Training Data for Enforcing Manufacturing Constraints on Topology Optimization}}, in: International Design Engineering Technical Conferences and Computers and Information in Engineering Conference, Vol. Volume 11A: 46th Design Automation Conference (DAC), 2020, p. V11AT11A005.
\newblock \href {https://doi.org/10.1115/DETC2020-22399} {\path{doi:10.1115/DETC2020-22399}}.
\newline\urlprefix\url{https://doi.org/10.1115/DETC2020-22399}

\bibitem{Ji2022}
S.~Ji, J.~Zhu, Y.~Yang, H.~Zhang, Z.~Zhang, Z.~Xia, Z.~Zhang, \href{http://dx.doi.org/10.3390/mi13060847}{Self-attention-augmented generative adversarial networks for data-driven modeling of nanoscale coating manufacturing}, Micromachines 13~(6) (2022) 847.
\newblock \href {https://doi.org/10.3390/mi13060847} {\path{doi:10.3390/mi13060847}}.
\newline\urlprefix\url{http://dx.doi.org/10.3390/mi13060847}

\bibitem{Killgore2023}
J.~P. Killgore, T.~J. Kolibaba, B.~W. Caplins, C.~I. Higgins, J.~D. Rezac, \href{http://dx.doi.org/10.1002/smll.202301987}{A data‐driven approach to complex voxel predictions in grayscale digital light processing additive manufacturing using u‐nets and generative adversarial networks}, Small 19~(50) (Jul. 2023).
\newblock \href {https://doi.org/10.1002/smll.202301987} {\path{doi:10.1002/smll.202301987}}.
\newline\urlprefix\url{http://dx.doi.org/10.1002/smll.202301987}

\bibitem{9627928}
D.~Wu, K.~Hur, Z.~Xiao, A gan-enhanced ensemble model for energy consumption forecasting in large commercial buildings, IEEE Access 9 (2021) 158820--158830.
\newblock \href {https://doi.org/10.1109/ACCESS.2021.3131185} {\path{doi:10.1109/ACCESS.2021.3131185}}.

\bibitem{Wang2019}
G.~Wang, A.~Ledwoch, R.~M. Hasani, R.~Grosu, A.~Brintrup, \href{http://dx.doi.org/10.1016/j.asoc.2019.105683}{A generative neural network model for the quality prediction of work in progress products}, Applied Soft Computing 85 (2019) 105683.
\newblock \href {https://doi.org/10.1016/j.asoc.2019.105683} {\path{doi:10.1016/j.asoc.2019.105683}}.
\newline\urlprefix\url{http://dx.doi.org/10.1016/j.asoc.2019.105683}

\bibitem{9388687}
C.~Liu, D.~Tang, H.~Zhu, Q.~Nie, A novel predictive maintenance method based on deep adversarial learning in the intelligent manufacturing system, IEEE Access 9 (2021) 49557--49575.
\newblock \href {https://doi.org/10.1109/ACCESS.2021.3069256} {\path{doi:10.1109/ACCESS.2021.3069256}}.

\bibitem{Shah2022}
M.~Shah, V.~Vakharia, R.~Chaudhari, J.~Vora, D.~Y. Pimenov, K.~Giasin, \href{http://dx.doi.org/10.1007/s00170-022-09356-0}{Tool wear prediction in face milling of stainless steel using singular generative adversarial network and lstm deep learning models}, The International Journal of Advanced Manufacturing Technology 121~(1–2) (2022) 723–736.
\newblock \href {https://doi.org/10.1007/s00170-022-09356-0} {\path{doi:10.1007/s00170-022-09356-0}}.
\newline\urlprefix\url{http://dx.doi.org/10.1007/s00170-022-09356-0}

\bibitem{harfoush2023framework}
A.~Harfoush, A.~Tabei, K.~R. Haapala, I.~Ghamarian, A framework for predicting grain morphology during incremental sheet metal forming using generative adversarial networks, Manufacturing Letters 35 (2023) 1081--1088, 51st SME North American Manufacturing Research Conference (NAMRC 51).
\newblock \href {https://doi.org/https://doi.org/10.1016/j.mfglet.2023.08.083} {\path{doi:https://doi.org/10.1016/j.mfglet.2023.08.083}}.

\bibitem{8715283}
S.~R. Chhetri, A.~B. Lopez, J.~Wan, M.~A. Al~Faruque, Gan-sec: Generative adversarial network modeling for the security analysis of cyber-physical production systems, in: 2019 Design, Automation and Test in Europe Conference and Exhibition (DATE), 2019, pp. 770--775.
\newblock \href {https://doi.org/10.23919/DATE.2019.8715283} {\path{doi:10.23919/DATE.2019.8715283}}.

\bibitem{zhou2022fedformer}
T.~Zhou, Z.~Ma, Q.~Wen, X.~Wang, L.~Sun, R.~Jin, Fedformer: Frequency enhanced decomposed transformer for long-term series forecasting, in: International conference on machine learning, PMLR, 2022, pp. 27268--27286.

\bibitem{patchtst}
Y.~Nie, N.~H. Nguyen, P.~Sinthong, J.~Kalagnanam, \href{https://arxiv.org/abs/2211.14730}{A time series is worth 64 words: Long-term forecasting with transformers} (2022).
\newblock \href {https://doi.org/10.48550/ARXIV.2211.14730} {\path{doi:10.48550/ARXIV.2211.14730}}.
\newline\urlprefix\url{https://arxiv.org/abs/2211.14730}

\end{thebibliography}

\end{document}